\documentclass[10pt,twocolumn,letterpaper]{article}

\usepackage{wacv}              

\usepackage{colortbl}
\usepackage{amsmath}
\usepackage{amsfonts}
\usepackage{algpseudocode}
\usepackage{algorithm}
\usepackage{booktabs} 

\usepackage{algpseudocode}
\usepackage{algorithmicx}
\usepackage{multirow}
\usepackage{fancyvrb}
\usepackage{arydshln}

\usepackage{graphicx}
\usepackage{subcaption}
\usepackage{wrapfig,lipsum,booktabs}
\usepackage{float}

\DeclareMathOperator*{\argmax}{arg\,max}
\DeclareMathOperator*{\argmin}{arg\,min}

\usepackage{tcolorbox}

\usepackage{amsthm}
\theoremstyle{plain}

\newtheorem*{thm*}{Theorem}
\theoremstyle{definition}

\newtheorem{cond}{Condition}

\newtheorem{case}{Case}

\newcommand{\Mat}{MAT}

\newcommand{\AugMatName}{Multimodal Augmented Adversarial Training}
\newcommand{\AugMatAbbrev}{MAT+}

\newcommand{\rowcolorbase}{\rowcolor{gray!20}}
\newcommand{\arrowup}[1]{\scriptsize{\textcolor{blue}{$\uparrow$#1}}} 
\newcommand{\arrowdown}[1]{\scriptsize{\textcolor{gray}{$\downarrow$#1}}}

\usepackage[accsupp]{axessibility}

\definecolor{wacvblue}{rgb}{0.21,0.49,0.74}
\usepackage[pagebackref,breaklinks,colorlinks,allcolors=wacvblue]{hyperref}

\def\wacvPaperID{711} 
\def\confName{WACV}
\def\confYear{2026}

\title{Multimodal Adversarial Defense for Vision-Language Models\\ by Leveraging One-To-Many Relationships}

\author{
Futa Waseda$^{1,3,\dag}$ \quad Antonio Tejero-de-Pablos$^{2}$ \quad Isao Echizen$^{1,3}$\\
$^1$The University of Tokyo \quad $^{2}$CyberAgent \quad $^3$National Institute of Informatics \\
{\tt\small $^{\dag}$futa-waseda@g.ecc.u-tokyo.ac.jp}
}

\begin{document}
\maketitle
\begin{abstract}
Pre-trained vision-language (VL) models are highly vulnerable to adversarial attacks. 
However, existing defense methods primarily focus on image classification, overlooking two key aspects of VL tasks: multimodal attacks, where both image and text can be perturbed, and the one-to-many relationship of images and texts, where a single image can correspond to multiple textual descriptions and vice versa (1:N and N:1). 
This work is the first to explore defense strategies against multimodal attacks in VL tasks, whereas prior VL defense methods focus on vision robustness.
We propose multimodal adversarial training (MAT), which incorporates adversarial perturbations in both image and text modalities during training, significantly outperforming existing unimodal defenses.
Furthermore, we discover that MAT is limited by deterministic one-to-one (1:1) image-text pairs in VL training data.
To address this, we conduct a comprehensive study on leveraging one-to-many relationships to enhance robustness, investigating diverse augmentation techniques.
Our analysis shows that, for a more effective defense, augmented image-text pairs should be well-aligned, diverse, yet avoid distribution shift---conditions overlooked by prior research.
This work pioneers defense strategies against multimodal attacks, providing insights for building robust VLMs from both optimization and data perspectives.
Our code is publicly available at \url{https://github.com/CyberAgentAILab/multimodal-adversarial-training}.

\end{abstract}

\vspace{-5pt}

\section{Introduction}
\label{sec:intro}

\begin{figure*}[t]
\vspace{-5pt}
    \centering
    \includegraphics[width=0.9\textwidth]{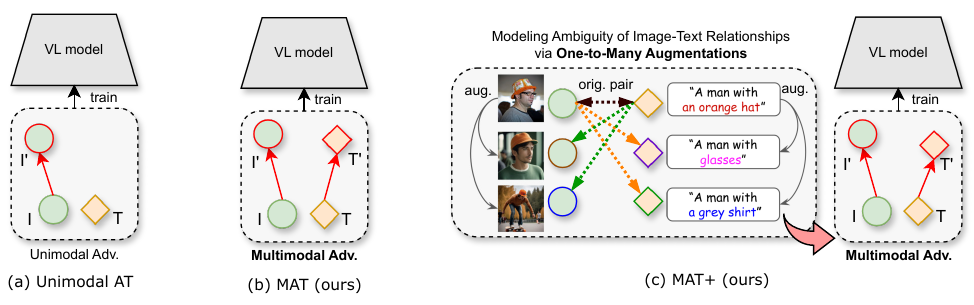}
    \vspace{-10pt}
    \caption{\textbf{Comparison of adversarial training (AT) methods for robust VL models.} (a) \textbf{Unimodal AT}, such as TeCoA~\cite{mao2022understanding} and FARE~\cite{schlarmann2024robust}, robustifies a single modality via unimodal adversarial examples (AEs). However, it overlooks two key aspects of VL tasks: \textit{multimodal attacks}, where attackers perturb both modalities, and \textit{one-to-many cross-modal alignment}, where an image has multiple valid descriptions, and vice versa. (b) \textbf{\Mat{}} addresses multimodal attacks by generating multimodal AEs during AT. (c) \textbf{\AugMatAbbrev{}} further captures the inherent ambiguity in image-text relationships via one-to-many augmentations.}
    \label{fig:teaser}
\vspace{-10pt}
\end{figure*}

Vision-language (VL) tasks require modeling the relationships between images and texts; for example, image-text retrieval (ITR) retrieves the most relevant text given an image query, and vice versa. 
Recent VL models such as CLIP~\citep{radford2021learning} achieve strong performance in these tasks; however, recent studies revealed that they are vulnerable to adversarial attacks~\citep{zhang2022towards, lu2023set}, which exploit nearly imperceptible input perturbations.
Such vulnerabilities pose serious practical risks; for instance, attackers may manipulate images or descriptions to alter retrieval rankings, unfairly promoting or demoting entities in recommendation systems.
As VL models are increasingly deployed, understanding and mitigating their adversarial vulnerabilities is urgent.

However, existing defense strategies for VL models~\citep{mao2022understanding, wang2024pre, schlarmann2024robust} primarily focus on image attacks (e.g., robust zero-shot image classification), leaving other VL tasks unexplored.
This is a considerable oversight for two reasons:
(1) \textbf{\textit{Multimodal manipulation}}: Attackers can perturb both images and texts, requiring more complex defense strategies than image-only methods.
(2) \textbf{\textit{One-to-many (1:N)} cross-modal alignment}: Unlike classification with simple and deterministic labels (e.g., ``a photo of a \{\textit{class}\}"), VL tasks involve diverse, ambiguous sentences (e.g., ``a man with glasses'' vs. ``a man wears an orange hat and glasses''), making robust image-text alignment more complicated.
By overlooking these aspects, existing defenses are limited in their effectiveness beyond zero-shot image classification.

To address this gap, we pioneer defense strategies for VL models against multimodal attacks.
Specifically, we study how to robustly fine-tune VL models for downstream VL tasks from both optimization and data perspectives, through extensive analysis. 

First, we propose \textbf{M}ultimodal \textbf{A}dversarial \textbf{T}raining (\Mat), which incorporates both image-text perturbations during training. Perturbing both modalities is non-trivial due to (1) the difficulty of simultaneous updates, (2) multiple objective choices for image-text attacks, and (3) increased computational complexity. 
Through extensive analysis, we designed \Mat{} to be both effective and reasonably efficient, while also offering valuable insights for future work.
\Mat{} significantly enhances multimodal robustness, demonstrating the necessity of defense methods tailored for multimodal attacks, an aspect overlooked by vision-only unimodal strategies~\citep{mao2022understanding, schlarmann2024robust}.

Furthermore, we argue that \Mat{}'s performance is limited by the inadequate approximation of the real data distribution when using the deterministic (1:1) image-text pairs contained in the training data.
To address this, we leverage the inherent one-to-many (1:N) relationships in VL data to enhance robustness.
Inspired by works in cross-modal ambiguity modeling in ITR~\citep{kim2023improving, song2019polysemous}, we explore augmentation techniques to create diverse one-to-many (1:N) and many-to-one (N:1) image-text pairs.
Our in-depth analysis reveals that augmentations are effective when pairs are well-aligned and diverse, without inducing distribution shift.
Specifically, text augmentations outperform image augmentations, since the higher dimensionality of images makes distribution shift harder to avoid.
Moreover, cross-modal augmentations (e.g., $image \rightarrow text$) outperform intra-modal ones (e.g., $text \rightarrow text$) by generating better-aligned pairs.
These findings provide novel insights into multimodal robustness and complement the literature on unimodal adversarial training.

Our contributions are summarized as follows:
\begin{itemize}
    \item \textbf{First defense strategy against multimodal attacks in VL models:} 
    We show that existing image-only defense methods are suboptimal for robust VL tasks and pioneer research in this new direction. 
    Specifically, we investigate strategies from both optimization and data perspective.
    \item \textbf{Proposed Multimodal Adversarial Training (\Mat):} 
    We designed \Mat{} to be both effective and efficient, through extensive analysis.
    \Mat{} largely improves multimodal robustness, highlighting the importance of considering multimodal perturbations in VL data.
    \item \textbf{Leveraging one-to-many relationships for robust VL tasks:} 
    We leverage one-to-many (1:N) image-text relationships via augmentations to enhance robustness, an aspect overlooked in unimodal adversarial training, which assumes a deterministic image-to-label mapping.
\end{itemize}

\vspace{-5pt}
\section{Related work}
\vspace{-5pt}



\textbf{Adversarial attacks on vision-language models.}
Adversarial attacks on VL models are categorized into unimodal and multimodal. 
Unimodal attacks, such as gradient-based image attacks~\citep{madry2017towards} and BERT-Attack for text~\citep{li2020bert}, perturb a single modality to mislead the models.
In contrast, multimodal attacks, which perturb both image and text modalities, are significantly more effective~\citep{zhang2022towards, lu2023set, han2023ot, wang2024transferable}.
However, developing defense strategies against multimodal attacks for VL tasks remains largely unexplored.

\textbf{Adversarial defense for vision-language models.}
Existing defense strategies for VL models mainly focus on vision robustness, in which adversarial attacks perturb only the image.
For example, \citet{mao2022understanding} and \citet{wang2024pre} approached zero-shot image classification on CLIP by proposing robust fine-tuning methods, which leverage unimodal adversarial training schemes to improve robustness.
\citet{schlarmann2024robust} also focused on image attacks only, and proposed a method for fine-tuning CLIP's vision encoder to improve robustness in several VL tasks (\ie, image classification, image-text retrieval).
Unlike the previous work, ours is the first to investigate adversarial defense strategies against multimodal attacks.
Specifically, we propose a multimodal adversarial training strategy to enhance robustness against such attacks in VL models and tasks. Furthermore, we leverage the one-to-many (1:N) image-text relationships to further improve adversarial robustness.

\textbf{Leveraging the one-to-many (1:N) nature of image-text.}
Recent works aimed at modeling the ambiguity between image and text pairs; a sentence may have multiple visual interpretations and an image may be described in various ways, however, typically only one pair is used as ground truth. Such deterministic 1:1 pairing is inconsistent with the natural 1:N relationships in the data.  
To address this, prior studies propose representing image-text samples as probabilistic embeddings~\citep{chun2021probabilistic,chun2024improved}, incorporating neighboring samples in the triplet loss~\citep{thomas2020preserving}, and generating multiple diverse representations for each pair~\citep{song2019polysemous,kim2023improving}. 
Inspired by these works, we leverage 1:N relationships to augment data in adversarial training and enhance VL robustness.

\section{Preliminaries}
\label{sec:preliminaries}

\textbf{Vision-language models.}
Many recent VL models (e.g., ALBEF~\cite{li2021align} and BLIP~\cite{li2023blip}) are fundamentally built on CLIP, which learns joint image-text representations via large-scale image-text contrastive learning. 
CLIP consists of an image encoder $f_{\theta_{\text{I}}}: \mathbb{R}^{d_{\text{I}}} \to \mathbb{R}^{d_\text{E}}$ and a text encoder $f_{\theta_{\text{T}}}: \mathbb{R}^{d_{\text{T}}} \to \mathbb{R}^{d_{E}}$, where $\theta_\text{I}$ and $\theta_\text{T}$ are their parameters, $d_{\text{I}}$ and $d_{\text{T}}$ are input dimensions, and $d_\text{E}$ is the joint embedding dimension.
Given an image $I \in \mathbb{R}^{d_{\text{I}}}$ and a text $T \in \mathbb{R}^{d_{\text{T}}}$, CLIP is trained to map them into a shared embedding space, maximizing the cosine similarity of image-text embeddings $S_{\theta_{\text{I},\text{T}}}(I, T) = \cos(f_{\theta_{\text{I}}}(I), f_{\theta_{\text{T}}}(T))$ for correct image-text pairs while minimizing it for incorrect pairs.
CLIP optimizes the InfoNCE loss for a batch of $N$ image-text pairs $\{(I_i, T_i)\}_{i=1}^N$ as:
\begin{equation}
    \label{eq:clip_loss}
    \mathcal{L}_{\text{CLIP-I}}(I, T) = - \sum_{i=1}^N \log \frac{\exp(S_{\theta_{\text{I},\text{T}}}(I_i,T_i)/\tau)}{\Sigma_{j=1}^N \exp(S_{\theta_{\text{I},\text{T}}}(I_i,T_j)/\tau)},
\end{equation}
where $\tau$ is the learnable temperature parameter. The overall loss is the average of the losses over images and texts, given by $\mathcal{L}_{\text{CLIP}} = (\mathcal{L}_{\text{CLIP-I}} + \mathcal{L}_{\text{CLIP-T}}) / 2$, where $\mathcal{L}_{\text{CLIP-T}}$ is the InfoNCE loss over texts.

\textbf{Multimodal adversarial attacks.}
We aim to defend VL models against adversarial attacks, where both image and text modalities are perturbed.
The objective of (untargeted) adversarial attacks on CLIP is to minimize the image-text similarity $S_{\theta_{I,T}}(I, T)$ for the correct image-text pairs $(I, T)$ to mislead the models' predictions as:
\begin{align}
    (\delta_I^\star, \delta_T^\star) 
&= \argmin_{\delta_I \in \Delta_I, \, \delta_T \in \Delta_T} 
    S_{\theta_{I,T}}\big(I + \delta_I,\, T + \delta_T\big), \\
(I', T') 
&= \big(I + \delta_I^\star,\, T + \delta_T^\star\big).
\end{align}
where $\delta_\text{I}$ and $\delta_\text{T}$ are image and text perturbations, and $\Delta_\text{I}$ and $\Delta_\text{T}$ define the set of allowed image and text perturbations.
Image attacks maintain perceptual similarity between $I$ and $I'$ typically using the $L_p$-norm.
A common image attack strategy is projected gradient descent (PGD)~\citep{madry2017towards}, which iteratively updates $I'$ by taking a small step in the direction of the gradient.
Text attacks, such as BERT-Attack~\citep{li2020bert}, modify $N$ tokens in the text $T$ to maximize the divergence between $f_{\theta_\text{T}}(T)$ and $f_{\theta_\text{T}}(T')$.
Multimodal attacks perturb both image-text modalities to create $(I', T')$; Co-Attack~\citep{zhang2022towards} perturbs each modality sequentially, first perturbing the image and then the text; SGA~\citep{lu2023set} enhances Co-Attack by considering the set-level interaction between multiple images and texts.

\textbf{Adversarial training for image classification.} Adversarial training (AT)~\citep{madry2017towards} is a widely used defense against adversarial attacks, where a model is trained on adversarial examples. It solves a min-max optimization problem:
\begin{align}
    \label{eq:adv_training}
    \argmin_{\theta_\text{C}} \mathbb{E}_{(x,y) \sim \mathcal{D}} \left( \max_{\delta_\text{I} \in \Delta_\text{I}} \left[ f_{\theta_\text{C}}(x+\delta_\text{I}) \neq y \right] \right), 
\end{align}
where $\theta_C$ represents the parameters of image classifier $f_{\theta_C}$, and $(x,y)$ denotes an image and its corresponding class label drawn from the data distribution $\mathcal{D}$.

Based on this, to improve CLIP's adversarial robustness in zero-shot image classification, TeCoA~\citep{mao2022understanding} adversarially fine-tunes CLIP's image encoder by minimizing the contrastive loss between adversarial images and the text expression of the corresponding class (``a photo of \{*\}''):
\begin{align}
    \argmin_{\theta_\text{I}} \mathbb{E}_{(I,T) \sim \mathcal{D}} \left( \max_{\delta_\text{I} \in \Delta_\text{I}} \mathcal{L}_{\text{{CLIP-I}}}(I+\delta_\text{I}, T) \right).
\end{align}
However, TeCoA only defends against image attacks.

\section{Methodology}
\label{sec:method}

We propose \textbf{M}ultimodal \textbf{A}dversarial \textbf{T}raining (\Mat) by addressing the limitations of existing unimodal AT (Fig.~\ref{fig:teaser}).
First, we describe its practical optimization strategies via image-text contrastive learning applicable to VL models.
Then, we introduce a data-driven approach that leverages the one-to-many relationship between images and texts to enhance robustness by more accurately approximating the true multimodal data distribution.

\subsection{Problem setup}
\label{sec:multimodal_AT}

To defend against multimodal attacks, we introduce \Mat, which perturbs both images and texts during AT.
Following the theory of unimodal AT (Eq.~\ref{eq:adv_training}), we formulate \Mat{} objective as:
\begin{align}
    \label{eq:mat}
    \min_{\theta} \mathbb{E}_{(I,T) \sim \mathcal{D}^{\star}} \left( \max_{\substack{\delta_\text{I} \in \Delta_\text{I}, \\ \delta_\text{T} \in \Delta_\text{T}}} \mathcal{L}_{\text{VL}}(I+\delta_\text{I}, T+\delta_\text{T}) \right),
\end{align}
where $\mathcal{D}^{\star}$ represents the true data distribution, $\theta$ represents the parameters of a VL model, and
$\mathcal{L}_{\text{VL}}$ refers to the VL training loss (e.g., $\mathcal{L}_{\text{CLIP}}$ for CLIP).

\subsection{Practical multimodal min-max optimization}
\label{sec:optim-algo}

\subsubsection{Multimodal inner maximization}
Directly solving the inner-maximization in Eq.~\ref{eq:mat} is highly non-trivial due to the difficulty of updating both modalities simultaneously, and the increased computational cost.

To address these challenges, we emphasize that exact maximization is not always necessary; in fact, the basic PGD-AT defense method~\cite{madry2017towards} approximates maximization with first-order adversaries.
Our extensive analysis led us to design \Mat{} to be both effective and reasonably efficient, while also offering insights for future multimodal defenses.

Specifically, we mainly adopt two practical strategies: (1) \textit{Step-by-step perturbation:} we generate adversarial examples $(I', T')$ by sequentially perturbing the text and image modalities, and (2) \textit{Loss simplification:} replacing $\mathcal{L}_{\text{VL}}$ with effective yet efficient alternative.

\textbf{Adversarial text generation.}
First, we generate adversarial texts using a representative text attack, BERT-attack~\cite{li2020bert}. BERT-attack identifies critical words contributing to the loss and replaces them with semantically similar, grammatically correct alternatives to maximize the loss.
However, this discrete optimization process is computationally expensive, and directly maximizing $\mathcal{L}_{\text{VL}}$ over multiple image-text pairs in a batch by repeatedly replacing words would be prohibitively time-consuming.
To address this, we exploit the shared image-text alignment objective in VL models (e.g., CLIP, ALBEF, BLIP), and instead of direct loss maximization, we maximize the divergence between individual image-text embeddings as:
\begin{equation}
    \label{eq:multimodal_attack_text}
    T' = T + \argmax_{\delta_\text{T} \in \Delta_\text{T}}  - \frac{ f_{\theta_\text{I}}(I) \cdot f_{\theta_\text{T}}(T + \delta_T) }{ \| f_{\theta_\text{I}}(I) \| \| f_{\theta_\text{T}}(T + \delta_T) \| }.
\end{equation}

\textbf{Adversarial image generation.}
Next, given adversarial texts $T'$, we generate adversarial images $I'$ using the widely adopted PGD attack~\cite{madry2017towards}.
Unlike the discrete optimization in text attacks, image attacks involve continuous optimization, where each update step requires a single back-propagation to maximize a loss function.
Therefore, directly maximizing the loss $\mathcal{L}_{\text{VL}}$ is practically feasible.
For CLIP, instead of maximizing $\mathcal{L}_{\text{CLIP}}$, we adopt a simple approach by minimizing the cosine similarity between the image-text embeddings as:
\begin{equation}
    \label{eq:multimodal_attack_image}
    I' = I + \argmax_{\delta_\text{I} \in \Delta_\text{I}} - \frac{ f_{\theta_\text{I}}(I+\delta_\text{I}) \cdot f_{\theta_\text{T}}(T') }{ \| f_{\theta_\text{I}}(I+\delta_\text{I}) \| \| f_{\theta_\text{T}}(T') \| }.
\end{equation}

For ALBEF and BLIP, we directly maximize their downstream objective functions, as they involve advanced techniques that cannot be trivially simplified.
Please see Appendix~\ref{appendix:mat_obj} for details.

\subsubsection{Multimodal outer minimization}
To update the model parameters, we minimize the train loss function $\mathcal{L}_{\text{VL}}$ using the generated adversarial images and texts.
For CLIP, we minimize the InfoNCE loss between the adversarial image $I'$ and text $T'$: $\min_{\theta} \mathcal{L}_{\text{CLIP}}(I', T')$.
For ALBEF and BLIP, we fine-tune the models with their original downstream-task-specific objectives.

\subsection{Leveraging one-to-many relationships for robustness generalization}
\label{sec:one-to-many}

While \Mat{} addresses the multimodality of VL tasks, the robustness can be further improved by leveraging also the one-to-many nature of image-text data. Thus, we propose \AugMatName{} (\AugMatAbbrev), which incorporates one-to-many augmentations to better approximate the true multimodal distribution of images and texts. 
Notably, this data-driven strategy for enhancing robustness is applicable to any VL model trained on image-text pairs.

\subsubsection{Approximating the multimodal distribution by modeling ambiguity via augmentations}
The true distribution of the data $\mathcal{D}^{\star}$ is inaccessible, but the dataset $\mathcal{\hat{D}}$ serves as an approximation, and its quality affects robustness significantly.
In the field of image classification, \citet{gowal2021improving} proved that adding image augmentations to the training data $\mathcal{\hat D}_\text{tr}$ produces a $\mathcal{\hat{D}}$ closer to $\mathcal{D}^{\star}$. This improves robustness under the assumption of a deterministic closed-set image-to-label mapping. However, their theory does not consider the ambiguity of VL data, where a single image can have multiple valid descriptions and vice versa (Fig.~\ref{fig:teaser}). Thus, for VL data, multimodal AT with (1:1) image-text pairs only provides a limited $\mathcal{\hat{D}}$.
To model this ambiguity, we generate diverse one-to-many (1:N) and many-to-one (N:1) image-text pairs through augmentations, improving the approximation of Eq.~\ref{eq:mat}.




\textbf{Modeling the ambiguity of text descriptions.}
Let $\mathcal{\hat D}_\text{tr}$ be a dataset of deterministic (1:1) image-text pairs $(I,T)$.
Relying solely on these pairs produces a weak $\mathcal{\hat D}$ because the original text $T$ often provides only a partial or subjective depiction of the image $I$ (Fig.~\ref{fig:teaser}).
In other words, $\mathcal{\hat D}_\text{tr}$ approximates the true image-to-text mapping $g_\text{T}^\star: \mathbb{R}^{d_\text{I}} \rightarrow \mathbb{R}^{d_\text{T}}$, a ``perfect'' human annotator capable of generating detailed descriptions.
However, striving for a true $g_\text{T}^\star$ is neither desirable nor practical; language compresses visual information by focusing on human interests, naturally involving ambiguity. Achieving $g_\text{T}^\star$ would require pixel-level descriptions, which are linguistically unrealistic.

Thus, instead of relying on a single caption per image, we embrace text ambiguity by generating one-to-many (1:N) image-text pairs.
In practice, given $\mathcal{\hat D}_\text{tr}$ with ground-truth pairs $(I, T)$, we can generate new aligned texts using $I$, $T$, or both. Consequently, Eq.~\ref{eq:mat} is approximated as:
\vspace{-1mm}
\begin{align}
    \label{eq:mat_approx_text}
    \min_{\theta} \mathbb{E}_{\substack{(I,T) \sim \mathcal{\hat D}_\text{tr}, \\ \phi \sim \Phi}} \left( \max_{\substack{\delta_\text{I} \in \Delta_\text{I}, \\ \delta_\text{T} \in \Delta_\text{T}}} \mathcal{L}_{\text{VL}}(I+\delta_\text{I},  {\hat g}_{\text{T},\phi}(I,T)+\delta_\text{T}) \right),
\end{align}
where ${\hat g}_{\text{T},\phi}$ is the approximated image-to-text model, with $\phi$ capturing the inherent randomness (ambiguity) in generation, and $\Phi$ is the set of all possible random variables.

\textbf{Modeling ambiguity of images.}
Similarly, a text description can correspond to multiple images, given the high dimensionality and degree-of-freedom of the image data space.
For example, the description ``a man with an orange hat'' could be simultaneously paired with various images, such as a man with a green shirt or engaging in different activities (Fig.~\ref{fig:teaser}).

Thus, as in the text case, we introduce image augmentations to create many-to-one image-text pairs to approximate $\mathcal{D}^\star$.
Formally, Eq.~\ref{eq:mat} is approximated as:
\vspace{-1mm}
\begin{align}
    \label{eq:mat_approx_prac_image}
    \min_{\theta} \mathbb{E}_{\substack{(I,T) \sim \mathcal{\hat D}_\text{tr}, \\ \psi \sim \Psi}} \left( \max_{\substack{\delta_\text{I} \in \Delta_\text{I}, \\ \delta_\text{T} \in \Delta_\text{T}}} \mathcal{L}_{\text{VL}}({\hat g}_{\text{I},\psi}(I,T)+\delta_\text{I},  T+\delta_\text{T}) \right), 
\end{align}
where ${\hat g}_{\text{I},\psi}$ is the approximated text-to-image model, with $\psi$ capturing the inherent randomness, and $\Psi$ denoting the set of all possible random variables.

\subsubsection{Conditions for effective augmentations}
\label{sec:cond_for_aug}
Not all $\Phi$ and $\Psi$ produce a model that can generate valid image-text pairs $\mathcal{I} \times \mathcal{T} \subseteq \mathbb{R}^{d_{\text{I}}} \times \mathbb{R}^{d_{\text{T}}}$, that is, paired data that effectively refine $\mathcal{\hat D}$ toward $\mathcal{D}^{\star}$.
We can develop the theory of \citet{gowal2021improving} for image classification to address multimodal VL augmentations. 
Let $p$ be the probability measure corresponding to $\mathcal{D}^{\star}$, for which every valid pair $(I,T) \in \mathcal{I} \times \mathcal{T}$ has non-zero probability: $p(I,T) > 0$. 
Similarly, let $\hat p$ be the probability measure corresponding to $\mathcal{\hat D}$.
The sufficient condition for $\mathcal{\hat D}$ to be an effective approximation is as follows:
\begin{cond}(Accurate approximation)
\label{cond:accurate-approx}
The approximated data distribution $\mathcal{\hat D}$ and true data distribution $\mathcal{D}^\star$ must be equivalent: $\forall (I,T) \in \mathcal{I} \times \mathcal{T}, p(I,T) = \hat p(I,T).$
\end{cond}



\textbf{Ineffective one-to-many augmentations.}
Let us denote augmentations for image-text pairs $(I,T) \in \mathcal{\hat D}_\text{tr}$ as $I_\text{aug}=\hat g_{\text{I},\psi}(I,T)$ and $T_\text{aug} = \hat g_{\text{T},\phi}(I,T)$.
Then, when adding an augmented pair $(I_\text{aug}, T)$ or $(I, T_\text{aug})$ to $\mathcal{\hat D}_\text{tr}$, we define three cases that result into ineffective augmentations, violating Cond.~\ref{cond:accurate-approx}.
\begin{case}(Semantic mismatch) 
\label{case:unmatch}
If augmented image-text pairs are not semantically aligned, $\hat p$ diverges from $p$ by generating data in regions of zero probability; that is, $\hat p(I_\text{aug},T)\neq p(I_\text{aug}, T) = 0$ or $\hat p(I,T_\text{aug})\neq p(I, T_\text{aug}) = 0$, meaning that the distributions $\mathcal{D}^\star$ and $\mathcal{\hat D}$ differ.
\end{case}
\begin{case}(Limited diversity) 
\label{case:lim_div}
Even if augmented image-text pairs are semantically aligned, trivial augmentations $I \simeq I_\text{aug}$ or $T \simeq T_\text{aug}$ do not provide enough variation to refine $\mathcal{\hat D}$ and capture the inherent ambiguity.
\end{case}
\begin{case}(Distribution shift) 
\label{case:dist-shift}
Even if augmented image-text pairs are aligned, excessive augmentations cause distribution shifts. This also applies to the unimodal case~\cite{gowal2021improving}.
\end{case}



Our results in Sec.~\ref{sec:results} show that \AugMatAbbrev{} with well-chosen augmentations avoiding these three cases significantly enhances robustness.

\section{Experimental settings}
\label{sec:analytical_setting}


We evaluate the multimodal robustness of defense methods on VLMs (i.e., CLIP and ALBEF) in their original tasks they were evaluated in, image-text retrieval (ITR) and visual grounding (VG). We also include BLIP, a VLM capable of text generation for image captioning (IC).

\begin{table*}[ht]
\vspace{-7pt}
    \centering
    \small
    \resizebox{1.0\textwidth}{!}{%
\begin{tabular}{lllcccccccccccccc}
\toprule
\multirow{3}{*}{\textbf{Method}} & \multirow{3}{*}{\textbf{Img aug.}} & \multirow{3}{*}{\textbf{Text aug.}} & \multicolumn{4}{c}{\textbf{Flickr30k}} & \multicolumn{4}{c}{\textbf{COCO}}  \\ \cmidrule(lr){4-7} \cmidrule(lr){8-11}
 & & & \multicolumn{2}{c}{\textbf{Clean}} & \multicolumn{2}{c}{\textbf{SGA}} &  \multicolumn{2}{c}{\textbf{Clean}} & \multicolumn{2}{c}{\textbf{SGA}} \\
\cmidrule(lr){4-5} \cmidrule(lr){6-7} \cmidrule(lr){8-9} \cmidrule(lr){10-11}
&  &  & TR@1 & IR@1 & TR@1 & IR@1 & TR@1 & IR@1 & TR@1 & IR@1 \\ \hline
\rowcolorbase Finetune &  &  & \textbf{92.1} & \textbf{77.2} & 0.6 & 0.6 & \textbf{66.6} & \textbf{50.1} & 0.1 & 0.1 \\
\hdashline \rowcolorbase FARE &  &  & 75.9 & 61.0 & 27.1 & 21.0 & 45.2 & 32.3 & 9.1 & 6.9 \\ 
\rowcolorbase TeCoA-ITR &  &  & 83.1 & 68.2 & 27.5 & 17.6 & 58.0 & 41.6 & 9.6 & 6.2 \\ \hline
(ours) \Mat &  &  & 84.6 & 67.7 & 36.4 & 24.9 & 55.8 & 40.7 & 17.7 & 12.3 \\
\hdashline (ours) \AugMatAbbrev &  & Basic(EDA) & 85.4 \arrowup{0.8} & 69.5 \arrowup{1.8} & 39.1 \arrowup{2.7} & 27.5 \arrowup{2.6} & 55.9\arrowup{0.2} & 40.2\arrowdown{0.5} & 18.4\arrowup{0.7} & 12.9\arrowup{0.6} \\
 &  & I2T(div-Caps) & 84.7 \arrowup{0.1} & 69.2 \arrowup{1.5} & 40.3 \arrowup{3.9} & 27.8 \arrowup{2.9} & 56.7\arrowup{0.9} & 39.9\arrowdown{0.8} & 18.9\arrowup{1.2} & 12.5\arrowup{0.2} \\
 &  & I2T(Human) & \underline{85.7} \arrowup{1.1} & \underline{71.9} \arrowup{4.2} & \textbf{45.6} \arrowup{9.2} & \textbf{32.2} \arrowup{7.3} & \underline{58.9}\arrowup{3.1} & \underline{43.1}\arrowup{2.4} & \textbf{21.3}\arrowup{3.6} & \textbf{14.5}\arrowup{2.2} \\
 & Basic(RandAug) &  & 84.1 \arrowup{0.5} & 67.1 \arrowdown{0.6} & 35.6 \arrowdown{0.8} & 24.4 \arrowdown{0.5} & 55.9\arrowup{0.1} & 40.7\arrowup{0.0} & 18.3\arrowup{0.6} & 12.6\arrowup{0.3} \\
 & T2I(SD) &  & 83.3 \arrowdown{1.3} & 68.4 \arrowup{0.7} & 37.9 \arrowup{1.5} & 25.2 \arrowup{0.3} & 54.2\arrowdown{1.6} & 38.3\arrowdown{2.4} & 17.2\arrowdown{0.5} & 11.7\arrowdown{0.6} \\
 & T-I2I(SD) &  & 83.8 \arrowup{0.8} & 69.1 \arrowup{1.4} & 39.3 \arrowup{2.9} & 25.8 \arrowup{0.9} & 57.1\arrowup{1.3} & 41.7\arrowup{1.0} & 18.8 \arrowup{1.1} & 12.6 \arrowup{0.2}
 \\
\bottomrule
\end{tabular}
    }
    \vspace{-7pt}
    \caption{
    \textbf{Comparison of CLIP trained on the Flickr30k and COCO datasets for image-text retrieval (ITR)} under the no-attack scenario (Clean) and multimodal attack (SGA), reporting R@k for text retrieval (TR@k) and image retrieval (IR@k).
    }
    \label{tab:clip_flickr_coco_itr}
\end{table*}

\begin{table*}[ht]
\vspace{-1pt}
    \centering
    \small
    \resizebox{1.0\textwidth}{!}{%
\begin{tabular}{lllcccccccccccccc}
\toprule
\multirow{3}{*}{\textbf{Method}} & \multirow{3}{*}{\textbf{Img aug.}} & \multirow{3}{*}{\textbf{Text aug.}} & \multicolumn{4}{c}{\textbf{Flickr30k}} & \multicolumn{4}{c}{\textbf{COCO}}  \\ \cmidrule(lr){4-7} \cmidrule(lr){8-11}
 & & & \multicolumn{2}{c}{\textbf{Clean}} & \multicolumn{2}{c}{\textbf{SGA}} &  \multicolumn{2}{c}{\textbf{Clean}} & \multicolumn{2}{c}{\textbf{SGA}} \\
\cmidrule(lr){4-5} \cmidrule(lr){6-7} \cmidrule(lr){8-9} \cmidrule(lr){10-11}
& & & TR@1 & IR@1 & TR@1 & IR@1 & TR@1 & IR@1 & TR@1 & IR@1 \\
\midrule
\rowcolorbase Finetune &  &  & \textbf{89.5} & \textbf{77.7} & 2.5 &  1.3 &\textbf{69.9} & \textbf{53.6} & 1.0  & 0.7 \\
\hdashline
\rowcolorbase TeCoA-ITR &  &  & 85.4 & 69.3  & 35.5 &  21.9  & 64.8 & 48.6 & 14.2 & 9.5 \\ \hline
(ours) \Mat &  &  & 82.0 & 66.3 & 47.1 &  32.9  & 63.9 & 46.2 & 31.2 & 21.2  \\
\hdashline
\multirow{7}{*}{(ours) \AugMatAbbrev} &  & Basic(EDA) & 82.2 \arrowup{0.2} & 67.9 \arrowup{1.6} &  44.6 \arrowdown{2.5}  & 31.2 \arrowdown{1.7} & 63.9 \arrowup{0.0} & 46.8 \arrowup{0.6} & 31.5 \arrowup{0.3} & 20.9 \arrowdown{0.3} \\
&  & I2T(div-Caps) & 85.6 \arrowup{3.6} & 71.0 \arrowup{4.8} & 48.8 \arrowup{1.7} & 35.0 \arrowup{2.1} & 66.0 \arrowup{2.0} & \underline{49.9} \arrowup{3.7}  & \underline{35.5} \arrowup{4.3} & 20.3 \arrowdown{0.9} \\
&  & I2T(Human) & \underline{85.8} \arrowup{3.8} & \underline{72.8} \arrowup{6.5} & \underline{52.9} \arrowup{5.8} & \textbf{38.8} \arrowup{5.9} & \underline{68.5} \arrowup{4.5} & 49.1 \arrowup{3.0} & \textbf{36.2} \arrowup{4.9} & \textbf{23.5} \arrowup{2.2} \\ 
& Basic(RandAug) &  & 82.2 \arrowup{0.2} & 67.2 \arrowup{0.9} & 48.3 \arrowup{1.2} & 33.4 \arrowup{0.5} &  63.1 \arrowdown{0.9} & 48.3 \arrowup{2.1}  & 30.3 \arrowdown{1.0} & 21.7 \arrowup{0.4} \\
& T2I(SD) &  & 83.7 \arrowup{1.7} & 68.3 \arrowup{2.1} & 52.0 \arrowup{4.9} &  36.2 \arrowup{3.3}  & 61.5 \arrowdown{2.4} & 46.0 \arrowdown{0.1} & 25.3 \arrowdown{5.9} & 18.6 \arrowdown{2.6} \\
& T-I2I(SD) &  & 85.1 \arrowup{3.1} & 69.8 \arrowup{3.6} & \textbf{55.2} \arrowup{8.1} & \underline{37.6} \arrowup{4.8} & 64.9 \arrowup{0.9} & 47.1 \arrowup{1.0} & 33.2 \arrowup{2.0} &  \underline{22.4} \arrowup{1.2} \\
\bottomrule
\end{tabular}
    }
    \vspace{-7pt}
    \caption{
    \textbf{Comparison of ALBEF trained on Flickr30k and COCO datasets for image-text retrieval (ITR)} under the no-attack scenario (Clean) and a multimodal attack (SGA), reporting R@k for text retrieval (TR@k) and image retrieval (IR@k).
    }
    \label{tab:albef_flickr_coco_itr}
\vspace{-9pt}
\end{table*}


\begin{table*}[]
\centering
\resizebox{1.0\textwidth}{!}{%
\begin{tabular}{lcccc:cc:cccc:cc:cc}
\toprule

 & \multicolumn{4}{c}{\textbf{Adversarial Training Config.}} & \multicolumn{2}{c}{{Clean}} & \multicolumn{2}{c}{{PGD (image)}} & \multicolumn{2}{c}{{BERT-Attack (text)}} & \multicolumn{2}{c}{\textbf{SGA (multimodal)}} & \multicolumn{2}{c}{\textbf{Time cost}} \\ \cmidrule(lr){2-5} \cmidrule(lr){6-7} \cmidrule(lr){8-9} \cmidrule(lr){10-11} \cmidrule(lr){12-13} \cmidrule(lr){14-15}
 & Order & \begin{tabular}[c]{@{}c@{}}Image Attack \\ (Obj., Optim.)\end{tabular} & \begin{tabular}[c]{@{}c@{}}Text Attack\\ (Obj., Optim.)\end{tabular} & \begin{tabular}[c]{@{}c@{}}Trained\\ params.\end{tabular} & IR@1 & TR@1 & IR@1 & TR@1 & IR@1 & TR@1 & IR@1 & TR@1 & (sec/iter) & (relative) 
 
 \\  \midrule

Finetune & - & - & - & All & 92.1 & 77.2 & 11.9 & 10.1 & 75.4 & 53.1 & 0.6 & 0.6 & 1.13 & $\times$ 0.13 \\ \hdashline
FARE & I & (Uni, PGD-10) & - & Vision & 75.9 & 61.0 & 69.7 & 55.1 & 53.2 & 40.2 & 27.1 & 21.0 & 7.81 & $\times$ 0.89 \\
TeCoA-ITR & I & (Cross, PGD-10) & - & All & 83.1 & 68.2 & 77.7 & 61.9 & 64.7 & 42.7 & 27.5 & 17.6 & 10.29 & $\times$ 1.17 \\ \hline \hline
\rowcolorbase MAT & T → I & (Cross, PGD-2) & (Cross, BERT) & All & 83.7 & 67.5 & 77.4 & 61.4 & 72.2 & 51.1 & 37.5 & 24.8 & 8.79 & - \\ \hdashline
\multicolumn{15}{l}{\textit{Objective ablation}} \\
\multicolumn{1}{c}{(1-1)} & T → I & (Uni, PGD-2) & (Cross, BERT) & All & 90.6 & 76.5 & 70.7 & 57.0 & 80.4 & 59.4 & 16.2 & 11.6 & 8.45 & $\times$ 0.96 \\
\multicolumn{1}{c}{(1-2)} & T → I & (Cross, PGD-2) & (Uni, BERT) & All & 83.4 & 66.8 & 79.3 & 64.2 & 71.0 & 49.4 & 33.3 & 22.8 & 8.74 & $\times$ 0.99 \\
\multicolumn{1}{c}{(1-3)} & T → I & (CLIP, PGD-2) & (CLIP, BERT) & All & -  & - & - & - & - & - & - & - & 460.7 & $\times$ 52.0 \\ \hdashline
\multicolumn{15}{l}{\textit{Perturbation order ablation}} \\
\multicolumn{1}{c}{(2-1)} & I → T & (Cross, PGD-2) & (Cross, BERT) & All & 84.4 & 68.7 & 80.5 & 65.0 & 74.8 & 51.2 & 36.3 & 24.6 & 8.79 & $\times$ 1.00 \\
\multicolumn{1}{c}{(2-2)} & T → I → T & (Cross, PGD-2) & (Cross, BERT) & All & 83.9 & 67.3 & 79.8 & 64.0 & 74.6 & 53.1 & 38.1 & 25.9 & 13.97 & $\times$ 1.59 \\ 
\multicolumn{1}{c}{(2-3)} & I → T → I & (Cross, PGD-2) & (Cross, BERT) & All & 82.3 & 65.7 & 79.9 & 62.7 & 69.3 & 49.1 & 37.4 & 25.1 & 11.10 & $\times$ 1.26 \\ \hdashline
\multicolumn{15}{l}{\textit{Perturbation strength ablation}} \\
\multicolumn{1}{c}{(3-1)} & T → I & (Cross, PGD-2) & EDA & All & 86.2 & 71.5 & 82.5 & 67.4 & 72.3 & 48.6 & 35.5 & 22.5 & 3.58 & $\times$ 0.41 \\
\multicolumn{1}{c}{(3-2)} & T → I & (Cross, PGD-10) & EDA & All & 84.3 & 67.8 & 80.9 & 65.7 & 68.4 & 45.7 & 36.0 & 23.7 & 12.18 & $\times$ 1.39 \\
\multicolumn{1}{c}{(3-3)} & T → I & (Cross, PGD-10) & (Cross, BERT) & All & 79.9 & 63.8 & 78.0 & 61.6 & 68.2 & 47.7 & 38.5 & 24.7 & 17.22 & $\times$ 1.96

\\
\bottomrule
\end{tabular}
}
\vspace{-5pt}
\caption{
\textbf{Analysis of \Mat's strategies.} Ablation study of CLIP trained on Flickr30k for image-text retrieval (ITR), comparing attack objectives (Cross- vs. Uni-modal), perturbation order (``Order''), and attack strength (PGD-2 vs. PGD-10, EDA vs. BERT).
}
\label{tab:ablation}
\vspace{-5pt}
\end{table*}

\begin{table}[ht]
    \centering
    \small
    \resizebox{0.9\columnwidth}{!}{%
\begin{tabular}{llccccccc}
\toprule
\multirow{2}{*}{\textbf{Method}} & \multirow{2}{*}{\textbf{Aug.}} & \multicolumn{2}{c}{\textbf{Clean}} & \multicolumn{2}{c}{\textbf{SGA}} \\ \cmidrule(lr){3-4} \cmidrule(lr){5-6}
& & TR@1 & IR@1 & TR@1 & IR@1 \\
\midrule
\rowcolorbase Finetune & & \textbf{72.9} & \textbf{57.5} & 1.2 & 1.1 \\ \hdashline
\rowcolorbase TeCoA-ITR &  & 64.6 & {51.8} &  20.2 & 13.9  \\ \hline
(ours) \Mat & & 66.9 & 49.9 & 31.3 & 21.0 \\ \hdashline
\multirow{2}{*}{(ours) \AugMatAbbrev} & I2T(Human) & \underline{71.0} \arrowup{4.1} & \underline{54.3} \arrowup{4.5} & \textbf{35.6} \arrowup{4.3} & \textbf{25.7} \arrowup{4.7} \\
 & T-I2I(SD) & 68.2 \arrowup{1.3} & 50.5 \arrowup{0.6}  & \underline{33.5} \arrowup{2.2} & \underline{22.9} \arrowup{1.9} \\
\bottomrule
\end{tabular}
    }
    \vspace{-5pt}
    \caption{
    \textbf{Comparison of BLIP trained on COCO for ITR} under the no-attack scenario (Clean) and the SGA attack, reporting R@k.
    }
    \label{tab:blip_coco_itr}
\vspace{-10pt}
\end{table}

\textbf{Datasets.}
We use the commonly used Flickr30k~\citep{plummer2015flickr30k} and the COCO~\citep{chen2015microsoft} datasets for ITR.
While these datasets contain five captions per image, our baseline training uses 1:1 image-text pairs (taking the first annotated caption), reflecting the typical setup when fine-tuning with real-world data, such as data collected from the internet, where 1:1 pairings are more common.
For VG, we use RefCOCO+~\cite{kazemzadeh2014referitgame}, and for IC, we use COCO. See Appendix~\ref{appendix:exp_setting_details} for details.

\textbf{Evaluation metrics.}
We use the recall@k (R@k) for evaluating ITR, including both image-to-text and text-to-image retrieval.
For VG we measure the accuracy in localizing the regions corresponding to the text descriptions, while IC evaluates the quality of generated captions using diverse metrics. Details are in Appendix~\ref{appendix:exp_setting_details}.

\textbf{Adversarial attacks.}
We evaluate defense methods against the multimodal adversarial attack SGA~\cite{lu2023set}, with perturbation constraints of $\epsilon_I = 2/255$ ($L_\infty$-norm) for images and one token for text, following \cite{zhang2022towards, lu2023set}.
Results on unimodal attacks (PGD, BERT-Attack) and Co-Attack~\cite{zhang2022towards} are in Appendix~\ref{abl:eval_other_results}.

\textbf{Baseline defense methods.}
Since there do not exist defense methods tailored for multimodal attacks, we compare our method with the unimodal (image-only) AT methods closest to our setting proposed for CLIP:
\begin{itemize}
\item \textit{FARE}~\cite{schlarmann2024robust}: An unsupervised (unimodal) adversarial fine-tuning scheme for CLIP, which focuses on obtaining a robust CLIP vision encoder. Since it is unsupervised, we can apply it to our ITR setting as-is.
\item \textit{TeCoA-ITR}: Since the original TeCoA~\cite{mao2022understanding} is text-guided, we extend it to be used in ITR. While the original TeCoA fine-tunes the vision encoder with image classification loss, TeCoA-ITR fine-tunes all parameters using cross-modal objective of Eq.~\ref{eq:multimodal_attack_image} to generate adversarial images.
\end{itemize}

\textbf{Training details.}
We fine-tune the pre-trained CLIP-ViT-B/16, ALBEF-14M, and BLIP w/ ViT-B models using \Mat. 
Adversarial images are generated via 2-step-PGD (perturbation size of 2/255 in $l_{\infty}$-norm), and adversarial texts using BERT-attack (1-token perturbation).
Please see the detailed hyperparameters in Appendix~\ref{appendix:hparam-details}.
Computational cost details are in Appendix~\ref{appendix:computational_cost}.

\textbf{Augmentation strategies.}
We consider two types of augmentations: \textit{intra-modal} and \textit{cross-modal}.
Intra-modal augmentation enhances data points without considering image-text interactions (text $\rightarrow$ text,  image $\rightarrow$ image), while cross-modal augmentation enhances data points by leveraging the other modality (image $\leftrightarrow$ text). Specifically, we explore the following augmentation techniques:
\begin{itemize}
\item Text augmentations:
    \begin{itemize}
        \item[$\circ$] \textit{Intra-modal}: EDA~\citep{wei2019eda} for basic word-level edits. LLM-based rewriting~\citep{fan2024improving} is shown in Appendix~\ref{abl:eval_other_results}. 
        \item[$\circ$] \textit{Cross-modal}: Image-to-text (I2T) generation using diverse prompts with InternVL~\citep{chen2024internvl} (``I2T(div-Caps)"), and human-generated captions from the remaining 4 annotations of Flickr30k and COCO (``I2T(Human)").
    \end{itemize}
\item Image augmentations:
    \begin{itemize}
        \item[$\circ$] \textit{Intra-modal}: RandAug~\citep{cubuk2020randaugment}, randomly applying affine transformations and color distortions.
        \item[$\circ$] \textit{Cross-modal}: Stable Diffusion (SD) for text-to-image  (T2I)~\citep{rombach2022high} and text-guided image-to-image (T-I2I)~\citep{meng2021sdedit}.
    \end{itemize}
\end{itemize}
Each original data point is augmented four times, leading to a $\times$5 expansion, except for T-I2I(SD), where using only two augmentations yielded better results.
Please see Appendix~\ref{appendix:details_augmentation} for the detailed augmentation settings.


\vspace{-2pt}
\section{Results}
\label{sec:results}

\vspace{-2pt}
\subsection{Image-text retrieval (ITR)}
\label{sec:res-itr}

\subsubsection{Effectiveness of Multimodal Adversarial Training}
\textbf{Defending against multimodal attacks requires multimodal perturbations.}
With multimodal adversarial perturbations, \Mat{} consistently achieves significantly greater robustness against multimodal attacks than the unimodal AT methods, FARE and TeCoA-ITR, which focus solely on image perturbations during AT.
The improvements are substantial and consistent for CLIP on Flickr30k and COCO (Tab.~\ref{tab:clip_flickr_coco_itr}), as well as ALBEF on both datasets (Tab.~\ref{tab:albef_flickr_coco_itr}).
These results highlight the necessity of defense strategies tailored for multimodal perturbations.


\textbf{Extensive analysis towards efficient and effective multimodal defense.}  
Solving the inner-maximization in Eq.~\ref{eq:mat} is non-trivial, since it requires updating both modalities and involves a high computational cost.
We conducted ablation studies on key design factors---objective functions, perturbation order, and perturbation strength---to make our defense efficient and effective (Tab.~\ref{tab:ablation}).

\begin{itemize}
    \item \textbf{Objective functions:} Crossmodal loss (cosine similarity between image-text pairs) outperforms unimodal loss (similarity between orig. and adv. samples), with image-side objective being particularly critical since the image modality is more vulnerable than text (MAT vs.\ (1-1)/(1-2)). For text perturbations, optimizing the full CLIP loss in BERT-Attack is prohibitively slow ((1-3)); disrupting a single image-text pair suffices with much lower cost.
    \item \textbf{Perturbation order:} The order of T$\rightarrow$I or I$\rightarrow$T has little effect, while T$\rightarrow$I introduced slightly better multimodal robustness (MAT vs.\ (2-1)). More sequences (e.g., T$\rightarrow$I$\rightarrow$T) slightly improves multimodal robustness but at a higher cost ((2-2), (2-3)).  
    \item \textbf{Perturbation strength:} \textit{Image:} Strength is controlled by PGD iterations. While FARE and TeCoA-ITR use a ten-steps PGD (PGD-10), \Mat{} achieves comparable efficiency by using PGD-2, sufficient for multimodal robustness. Using PGD-10 further improves robustness, at the cost of clean accuracy and efficiency ((3-3)). \textit{Text:} EDA (simple word edits) is an efficient alternative to BERT-Attack ((3-1), (3-2)), since single-token perturbations in the discrete space are largely covered by simple edits.
\end{itemize}

\subsubsection{Effectiveness of one-to-many augmentations}

\begin{figure}[t]
\vspace{-5pt}
    \centering
    \includegraphics[width=\columnwidth]{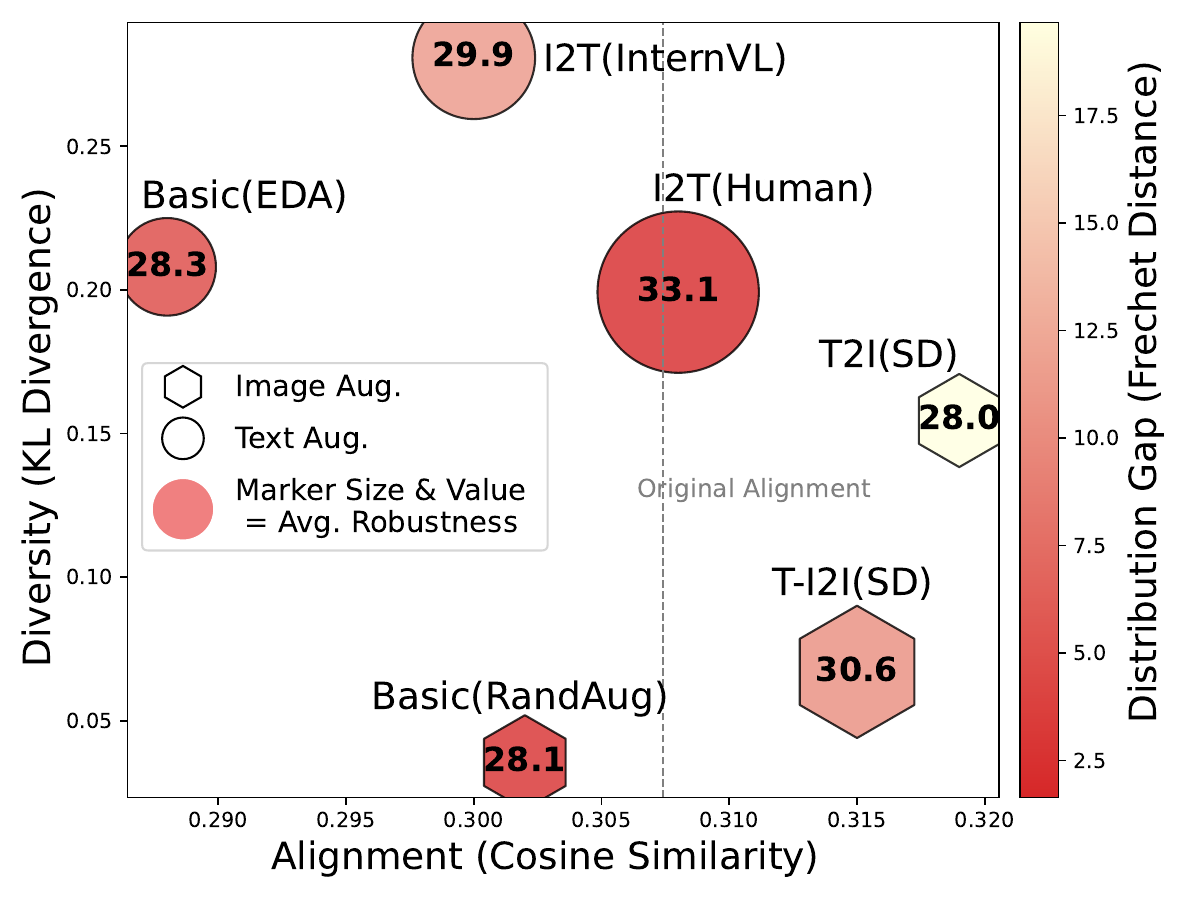}
    \vspace{-20pt}
    \caption{The relationships between the three properties of augmentations, (i) \textit{alignment}, (ii) \textit{diversity}, and (iii) \textit{distribution gap}, versus Robust Accuracy against multimodal attack (the overall average of IR@1/TR@1 for CLIP/ALBEF on Flickr/COCO). }
    \label{fig:alignment-diversity-distgap}
\vspace{-10pt}
\end{figure}

Through our comprehensive analysis, we reveal that some augmentations consistently enhance robustness, whereas some do not.
To identify key factors for effective augmentation, we analyze \textit{alignment}, \textit{diversity}, and \textit{distribution gap} based on our hypothesis for approximating the true image-text distribution (Sec.\ref{sec:cond_for_aug}).
To analyze this, we introduce three measurable properties:
\begin{itemize}
    \item \textbf{Alignment}: The semantic similarity between augmented image-text pairs: $S_{\theta_{\text{I},\text{T}}}(I_\text{aug}, T)$ or $S_{\theta_{\text{I},\text{T}}}(I, T_\text{aug})$.
    \item \textbf{Diversity}: The Kullback-Leibler (KL) divergence between the original and augmented samples: $d_\text{KL}(I, I_\text{aug})$ or $d_\text{KL}(T, T_\text{aug})$.
    \item \textbf{Distribution gap}: The Fr\'echet distance~\cite{frechet1936coefficient} between the distributions of the original and augmented samples, $d_\text{F}(\mathcal{\hat{D}}_\text{tr}, \mathcal{\hat{D}}_\text{aug})$, where $\mathcal{\hat{D}}_\text{aug}={(I_\text{aug}, T),(I, T_\text{aug})}^N_{i=1}$. This metric, widely used for evaluating generative models (e.g., FID~\cite{heusel2017gans}), quantifies the distribution gap.
\end{itemize}
High \textit{alignment} reduces semantic mismatches (Case~\ref{case:unmatch}), sufficient \textit{diversity} captures image-text ambiguity, preventing Case~\ref{case:lim_div}, and minimal \textit{distribution gap} avoids generating out-of-distribution samples, preventing Case~\ref{case:dist-shift}.

Fig.~\ref{fig:alignment-diversity-distgap} reveals that all properties, high alignment, high diversity, and small distribution gap, is crucial for effective augmentations.

\textbf{Cross-modal augmentation outperforms intra-modal augmentation due to higher image-text alignment.}
We observe that cross-modal augmentations yield better robustness than intra-modal ones, by considering the other modality and generating well-aligned image-text pairs.
For example, the alignment scores of RandAug and EDA are lower than those of the original pairs (Fig.~\ref{fig:alignment-diversity-distgap}), indicating \textit{semantic mismatch} (Case.~\ref{case:unmatch}).
Intra-modal augmentations struggle to balance diversity and alignment: increasing diversity without considering image-text relationships disrupts alignment. 
Fig.~\ref{fig:alignment-diversity-distgap} shows that intra-modal augmentations appear in the lower-left, achieving only one of alignment or diversity.

\textbf{Achieving sufficient diversity while keeping the distribution gap minimal is crucial.}
Figure~\ref{fig:alignment-diversity-distgap} indicates that even if augmentations are well-aligned as well as sufficiently diverse to expand the training data, they do not enhance robustness when there is a large distribution gap from the original samples.
For example, T2I(SD) introduces a good diversity but also introduces a large distribution gap, resulting in minimal robustness improvement.

Appendix~\ref{sec:vis_aug} provides qualitative examples of the augmentations generated for each technique.





\begin{figure}[t]
    \centering
    \vspace{-5pt}
    \begin{subfigure}{\columnwidth}
        \centering
        \includegraphics[width=0.85\textwidth]{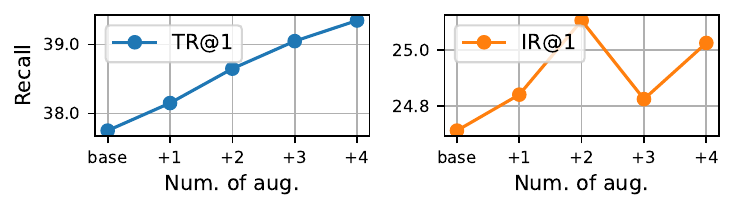}
        \vspace{-7pt}
        \caption{T-I2I(SD)}
        \label{fig:aug_num_sd}
    \end{subfigure}
    \begin{subfigure}{\columnwidth}
        \centering
        \includegraphics[width=0.85\textwidth]{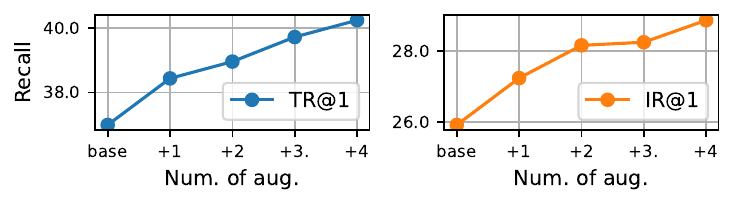}
        \vspace{-7pt}
        \caption{I2T(Human)}
        \label{fig:aug_num_caps}
    \end{subfigure}
    \vspace{-17pt}
    \caption{Analysis of the number of augmentations and robustness against SGA attacks for: (a) image augmentations using T-I2I(SD), and (b) text augmentations using I2T(Human).}
    \label{fig:analysis-aug-num}
\vspace{-10pt}
\end{figure}

\textbf{Image augmentations are challenging.}
While both image and text augmentations proved effective, text performs better.
This is because the high-dimensional image space makes it difficult to generate augmentations that are diverse yet distribution-consistent, whereas the structured nature of text allows controlled augmentation.  
Figure~\ref{fig:alignment-diversity-distgap} shows that text augmentations can introduce diversity while maintaining a small distribution gap, unlike image augmentations.
Additionally, Fig.~\ref{fig:analysis-aug-num} analyzes the number of augmentations.
Increasing text augmentations with I2T(Human) improves robustness, as more captions better approximate the data distribution, however, image augmentations using T-I2I(SD) enhance robustness up to two additional augmentations but can degrade it beyond that due to distribution shift.
These results suggest that developing image augmentations with high diversity yet minimal distribution gap is a promising direction for future research.

\textbf{Many-to-many augmentations (N:N).}
Appendix~\ref{abl:eval_other_results} shows that combining image and text augmentations (N:N) did not enhance performance over text-only augmentations. 
With two augmented images and four texts, original pairs constitute only a 12.5\% of the data, which makes (N:N) augmentations prone to distorting the data distribution if not designed carefully.
Thus, while theoretically promising, many-to-many augmentations require a dedicated methodology, which falls out of the scope of this work.

\begin{table}[t]
    \centering
    \small
    \resizebox{0.85\columnwidth}{!}{%
\begin{tabular}{lllccc}
\toprule
\multirow{2}{*}{\textbf{Method}} & \multirow{2}{*}{\textbf{Aug.}} & \multicolumn{3}{c}{\textbf{SGA}} \\  \cmidrule(lr){3-5}
 &  & Val &  Test-A &  Test-B  \\
\midrule
\rowcolorbase Fine-tune &  & 32.9 & 36.3 & 29.6 \\ \hdashline
\rowcolorbase TeCoA-ITR &  & 38.8 & 44.8 & 32.9 \\ \hline
(ours) \Mat &  & 40.4 & 44.2 & 35.1 \\ \hdashline
\multirow{3}{*}{(ours) \AugMatAbbrev} & I2T(div-Caps) & \textbf{43.2} \arrowup{2.8} & \textbf{48.1} \arrowup{3.9} & \textbf{37.1} \arrowup{2.1} \\
 & I2T(Human) & \underline{42.0} \arrowup{1.6} & \underline{46.5} \arrowup{2.3} & \underline{35.7} \arrowup{0.6} \\
  & T-I2I(SD$)$ & 41.1 \arrowup{0.7} & 46.0 \arrowup{1.8} & 34.8 \arrowdown{0.3} \\
\bottomrule
\end{tabular}
    }
    \vspace{-7pt}
    \caption{Accuracy comparison using ALBEF on the RefCOCO+ dataset for visual grounding.
    }
    \label{tab:albef_refcoco_vg}
    \vspace{-5pt}
\end{table}

\begin{table}[t]
    \centering
    \small
    \resizebox{0.85\columnwidth}{!}{%
\begin{tabular}{lllccc}
\toprule
\multirow{2}{*}{\textbf{Method}} & \multirow{2}{*}{\textbf{Aug.}} & \multicolumn{3}{c}{\textbf{SGA}} \\ \cmidrule(lr){3-5}
 &  &  BLEU-4 & ROUGE & CIDEr  \\
\midrule
\rowcolorbase Finetune &  & 11.1 & 35.6 & 35.5 \\ \hdashline
\rowcolorbase FARE & & \underline{23.6} & \underline{49.0} & \underline{77.8}  \\ \hline
(ours) \Mat  &  & \underline{21.8} & \underline{46.8} & \underline{74.4} \\ \hdashline
\multirow{1}{*}{(ours) \AugMatAbbrev} & I2T(Human) & \textbf{25.3} \arrowup{3.5} & \textbf{49.9} \arrowup{3.1} & \textbf{83.6} \arrowup{9.2} \\
\bottomrule
\end{tabular}
    }
    \vspace{-7pt}  
    \caption{Generated text quality comparison using BLIP evaluated on the COCO dataset for IC.
    }
    \label{tab:blip_image_captioning}
\vspace{-10pt}  
\end{table}

\textbf{Ablation study: MAT+ vs. naively increasing data samples.}
Appendix~\ref{appendix:fixed_data_size} (Tab.~\ref{tab:controlled_exp}) presents controlled experiments to disentangle the effect of data size from the one-to-many (1:N) strategy.
With the same number of data points, MAT+ outperforms the naive 1:1 settings.

\subsection{Additional tasks}

\textbf{Visual grounding (VG).}
We evaluate \Mat{} in VG using the model and dataset originally proposed for this task, ALBEF and RefCOCO+ (Tab.~\ref{tab:albef_refcoco_vg}). \Mat{} proves to be the most effective in this task as well, while \AugMatAbbrev{} further improves robustness.
This demonstrates the effectiveness of one-to-many augmentations in VL tasks beyond ITR.

\textbf{Image captioning (IC).}
We verify the effectiveness of one-to-many augmentations in IC using the model and dataset originally proposed for this task, BLIP and COCO (Tab.~\ref{tab:blip_image_captioning}).
Since IC involves only image inputs, \Mat{} does not surpass FARE, the unimodal AT method tailored for image attacks. Nevertheless, augmentations still enhance robustness in this task, with \AugMatAbbrev{} outperforming FARE.

\section{Conclusions}
This work is the first to study adversarial defense for vision-language (VL) models against multimodal attacks in VL tasks.
Existing defenses fail against multimodal attacks as they focus on image-only perturbations and overlook the one-to-many nature of image-text pairs.
To address this, we proposed \Mat{}, a novel defense framework that leverages multimodal perturbations and one-to-many augmentations.
Through comprehensive analysis, we design efficient yet effective multimodal defenses from an optimization perspective and introduce a data-driven approach to further enhance robustness. 
Our findings reveal key challenges, such as the difficulty in complex optimization and the need for diverse yet in-distribution augmentations, guiding future research.

\paragraph{Acknowledgment}
This work was partially supported by JSPS KAKENHI Grants JP21H04907 and JP24H00732, by JST CREST Grants JPMJCR20D3 and JPMJCR2562 including AIP challenge program, by JST AIP Acceleration Grant JPMJCR24U3, and by JST K Program Grant JPMJKP24C2 Japan.



{
    \small
    \bibliographystyle{ieeenat_fullname}
    \bibliography{main}
}

\clearpage 
\onecolumn
\appendix

\section{Experimental Details}
\label{appendix:exp_setting_details}

\subsection{\Mat's Inner Maximization Strategy}
\label{appendix:mat_obj}

\begin{table*}[ht]
\centering
\small
\begin{tabular}{lllll}
\toprule
\textbf{Model} & \textbf{VL Task} & \multicolumn{2}{c}{\textbf{Inner maximization}} & \textbf{Outer minimization} \\
& & \textbf{Text Adv. Generation} & \textbf{Image Adv. Generation} & \\ 
\midrule
\textbf{CLIP} & Image-Text Retrieval & $\max_{\delta_T} \cos( f_{\theta_T}(T+\delta_T), f_{\theta_I}(I)) $ & $\min_{\delta_I} \cos(f_{\theta_{\text{I}}}(I+\delta_I), f_{\theta_{\text{T}}}(T))$ & $\mathcal{L}_{\text{CLIP}}$  \\ \midrule
\textbf{ALBEF} & Image-Text Retrieval & $\max_{\delta_T} \cos( f_{\theta_T}(T+\delta_T), f_{\theta_I}(I)) $ & $\max_{\delta_I} \mathcal{L}_{\text{ITC}} + \mathcal{L}_{\text{ITM}} $ & $\mathcal{L}_{\text{ITC}} + \mathcal{L}_{\text{ITM}}$ \\
 & Visual Grounding & $\max_{\delta_T} \cos( f_{\theta_T}(T+\delta_T), f_{\theta_I}(I)) $ & $\max_{\delta_I} \mathcal{L}_{\text{ITC}} + \mathcal{L}_{\text{ITM}} $ & $\mathcal{L}_{\text{ITC}} + \mathcal{L}_{\text{ITM}}$\\ \hline
\textbf{BLIP} & Image-Text Retrieval & $\max_{\delta_T} \cos( f_{\theta_T}(T+\delta_T), f_{\theta_I}(I)) $ & $\max_{\delta_I} \mathcal{L}_{\text{ITC}} + \mathcal{L}_{\text{ITM}} $ & $\mathcal{L}_{\text{ITC}} + \mathcal{L}_{\text{ITM}}$ \\ 
& Image Captioning & $\max_{\delta_T} \cos( f_{\theta_T}(T+\delta_T), f_{\theta_I}(I)) $ & $\max_{\delta_I} \mathcal{L}_{\text{LM}} $ & $\mathcal{L}_{\text{LM}}$ \\ 
\bottomrule
\end{tabular}
\caption{Summary of \Mat's strategies for each model and downstream task.}
\label{tab:mat_obj}
\end{table*}

In the main text, we proposed \Mat{}, a multimodal defense framework and practical optimization strategies.
In our experiments, we adaptively modify \Mat's inner maximization strategy for different downstream tasks, as summarized in Table~\ref{tab:mat_obj}.
We provide detailed explanations below.

\subsubsection{Image-Text Retrieval (ITR)}

\Mat{} for CLIP generates text adversarial examples by maximizing the divergence between image-text embeddings, using BERT-attack~\cite{li2020bert}, while image adversarial examples are generated by minimizing their cosine similarity, following image attacks in Co-Attack~\cite{zhang2022towards} and SGA attack~\cite{lu2023set}. 

The ALBEF and BLIP model architectures use a similar approach for ITR but include additional components, such as an ITM module (Image-Text Matching) and multimodal encoders.
ITC loss ($\mathcal{L}_{\text{ITC}}$) is the image-text contrastive loss, similar to $\mathcal{L}_{\text{CLIP}}$, with the primary difference being whether momentum encoders are used to store previously seen representations. ITM loss ($\mathcal{L}_{\text{ITM}}$) predicts whether an image-text pair is positive (matched) or negative (not matched), which is predicted by a multimodal encoder in ALBEF and BLIP in addition to their unimodal image and text encoders. Please refer to original papers~\cite{li2021align, li2023blip} for more details.
Given the added complexity in ALBEF and BLIP, \Mat{} for these models is designed as follows:
\begin{itemize}
    \item Text attack: Same strategy as training CLIP with \Mat.
    \item Image attack: Directly maximize the downstream-task specific fine-tuning objective.
\end{itemize}

\subsubsection{Visual Grounding (VG)}
In our experiment, we evaluated the effectiveness of \Mat{} on visual grounding (VG) using ALBEF. Since the fine-tuning objective for VG is the same as for ITR, the inner maximization strategy of \Mat{} is identical to \Mat{} for ITR.

\subsubsection{Image Captioning}
In our experiment, we evaluated the effectiveness of \Mat{} on image captioning using BLIP. The fine-tuning objective for image captioning includes ITC loss and language modeling (LM) loss, and we maximize the sum of these objectives for perturbing images. 



\subsection{Evaluation Settings}
\label{appendix:eval_setting}

\subsubsection{Image-Text Retrieval (ITR)}
\noindent \textbf{Dataset.}
We use Flickr30k, which consists of image-text pairs with a train/test split of 29,000/1,000 images, and COCO, which consists of image-text pairs with a train/val/test split of 113,287/5,000/5,000 images.
We use the default split for training and testing.
While each image has five captions, our baseline training approach uses 1:1 image-text pairs, reflecting the practical setting for fine-tuning VL models.

\noindent \textbf{Evaluation metric.}
We use R@k, which measures the recall of the correct image or text among the top-k retrieved candidates. These metrics assess the model's ability to correctly retrieve relevant images or texts when given a query.

\subsubsection{Visual Grounding (VG)}

\noindent \textbf{Dataset.}
For training, we use MSCOCO's training set, following the ITR setting.
For testing VG tasks, we use RefCOCO+, which includes text descriptions of objects in images along with their corresponding bounding boxes.
RefCOCO+ contains 141,564 expressions for 19,992 images from the COCO training set.
This dataset was collected interactively through a two-player game~\citep{kazemzadeh2014referitgame}.
It consists of three test sets, \text{Val}, \text{TestA}, and \text{TestB}. Test Set A contains objects sampled randomly from the entire dataset. Test Set B contains objects sampled from the most frequently occurring object categories in the dataset. Test Set C contains objects sampled from images that contain at least 2 objects of the same category.

\noindent \textbf{Evaluation metric.}
VG aims to localize the region in an image that corresponds to a specific text description.
Following \citet{li2021align}, we extend Grad-CAM~\cite{selvaraju2017grad} to acquire heatmaps, and use them to rank the detected proposals provided by \citet{yu2018mattnet}.
We measure the accuracy of the attention maps with the IoU threshold being 0.5.

\subsubsection{Image Captioning}

\noindent \textbf{Dataset.}
We use the MSCOCO dataset for both training and testing. For training data, we follow the ITR setting.

\noindent \textbf{Evaluation metric.}
We use various evaluation metrics to measure the quality of the generated captions, including BLEU~\citep{papineni2002bleu}, METEOR~\cite{banerjee2005meteor}, ROUGE~\cite{lin2004automatic}, CIDEr~\cite{vedantam2015cider}, and SPICE~\cite{anderson2016spice}.

\subsection{Augmentation techniques}
\label{appendix:details_augmentation}

\subsubsection{Text augmentations}

\noindent \textbf{EDA (Easy Data Augmentation).}
EDA randomly selects words in the text and performs the following operations: synonym replacement, random insertion, random swap, or random deletion. We use the official implementation~\footnote{https://github.com/jasonwei20/eda\_nlp}.
The hyperparameter $\alpha$ controls the strength of the augmentation, where $\alpha$ determines the probability of each word being augmented.
We use $\alpha=0.3$ for all experiments.

\noindent \textbf{LangRW (Language rewrite).}
Language rewrite (LangRW)~\citep{fan2024improving} is a method that rewrites the text data to improve the robustness of the model, using a generative natural language processing model, such as Llama~\citep{touvron2023llama}.
We used Llama-2-7B~\footnote{https://huggingface.co/meta-llama/Llama-2-7b}.
In our work, we used slightly modified prompts from the original work to simultaneously generate four captions per image.
Given an original caption $T$, the prompt for generating additional captions are as follows:
\small
\begin{Verbatim}[frame=single,samepage=true]
Rewrite image captions in 4 different
ways.
    
{coco caption 1 for image i}
=> {coco caption 2 for image i}
=> {coco caption 3 for image i}
=> {coco caption 4 for image i}
=> {coco caption 5 for image i}

{coco caption 1 for image j}
=> {coco caption 2 for image j}
=> {coco caption 3 for image j}
=> {coco caption 4 for image j}
=> {coco caption 5 for image j}

{coco caption 1 for image k}
=> {coco caption 2 for image k}
=> {coco caption 3 for image k}
=> {coco caption 4 for image k}
=> {coco caption 5 for image k}

{original caption to be rewritten}
=>
\end{Verbatim}
\normalsize
where the coco captions are randomly sampled from the original captions from the COCO dataset~\citep{chen2015microsoft}.


\noindent \textbf{I2T(div-Caps) using InternVL.}
InternVL~\cite{chen2024internvl} is a latest vision-language multimodal model. We use the InternVL-2.5-2B model. We generate four captions per image using the following four prompts to ensure diversity:
\begin{itemize}
    \item \textit{Details}: ``Describe the image in detail.''
    \item \textit{MainObj}: ``Describe only the one main object in the image, do not say anything about the other objects or background.''
    \item \textit{Background}: ``Describe only the background of this image, do not say anything about the foreground objects.''
    \item \textit{Style}: ``Describe the style or your feelings about this image, do not say anything about the objects in the image.''
\end{itemize}
The lower alignment of InternVL-generated captions compared to original image-text pairs (Fig.~\ref{fig:alignment-diversity-distgap}) is due to the fact that captions generated with \textit{Background} and \textit{Style} are less aligned with images than the full captions.

\noindent \textbf{Human.} 
Human augmentation is a method that generates additional captions by human annotators.
Since we use 1:1 image-text pairs for training as default, we used the rest of the original captions included in Flickr30k and COCO datasets as additional captions for each image.

\subsubsection{Image augmentations}

\noindent \textbf{RandAug (Random Augmentation).}
RandAug~\citep{cubuk2020randaugment} is an image augmentation method that applies a series of random transformations to the image.
We used the codes from ALBEF~\citep{li2021align}~\footnote{https://github.com/salesforce/ALBEF}.
We set the number of operations to 2 and the magnitude to 5 for all experiments.

\noindent \textbf{Stable Diffusion (SD); Text-to-Image.}
We used SD-v2.1~\footnote{https://huggingface.co/stabilityai/stable-diffusion-2-1-base} for text-to-image augmentations, using Huggingface's default hyperparameters.

\noindent \textbf{Stable Diffusion (SD); Text-guided Image-to-Image.}
We used SD-v2.1, and use the pipeline for text-guided image-to-image generation~\cite{meng2021sdedit} from Huggingface\footnote{\url{https://huggingface.co/docs/diffusers/api/pipelines/stable_diffusion/img2img}}.
We used a strength hyperparameter of 0.5, which controls the diversity of augmentations.
Larger strength increases diversity, however, caused a distribution shift that negatively impacted performance.

\subsection{Hyperparameter Details}
\label{appendix:hparam-details}

We fine-tune the pre-trained CLIP-ViT-B/16, ALBEF-14M, and BLIP w/ ViT-B.
We generate adversarial images using 2-step-PGD with a step size of $1.0 / 255$ with the perturbation bound $\epsilon_\text{I} = 2.0 / 255$. 
For adversarial texts, we generate with BERT-attack, which replaces a word from top-k candidates with $k=10$.
We train for 5,000 steps on Flickr30k and 10,000 steps on COCO, with a batch size of 128.
For CLIP, we train using the SGD optimizer, while for ALBEF and BLIP, we use the AdamW optimizer. All models are trained with cosine learning rate scheduling, a learning rate of 0.0001, and a weight decay of 0.0001.

\subsection{Computational Cost}
\label{appendix:computational_cost}
All experiments were conducted on a single NVIDIA A100 GPU.

Since multimodal defense requires perturbing both image and text modalities, naively combining PGD and BERT-Attack nearly doubles the cost compared to unimodal adversarial training methods (e.g., FARE, TeCoA). More complex perturbation sequences (e.g., T$\!\rightarrow$I$\!\rightarrow$T) are also computationally prohibitive. 
To address this, we carefully designed \Mat{} to be both effective and efficient.

Tab.~\ref{tab:ablation} in the main text compares MAT's training time with that of different ablations of our method.
On the one hand, a naive setting with ten-steps PGD (PGD-10) and BERT-Attack (MAT (3-3)) nearly doubles the cost. On the other hand, since baseline unimodal defenses perturb only the image modality, their training time is faster.
In comparison, when \Mat{} adopts PGD-2 for image perturbations and a single step-by-step sequence (T$\!\rightarrow$I), we achieve an efficiency comparable to that of FARE and TeCoA-ITR while significantly enhancing multimodal robustness.  
Thanks to this, training time remains feasible in practice (e.g., CLIP: 12h, BLIP: 24h on COCO); this modest overhead compared to previous---unimodal---approaches is justified by the substantial robustness gains.

In \AugMatAbbrev{}, the cost of the augmentations depends on each technique. Cross-modal augmentations using generative models (e.g., InternVL for captioning, Stable Diffusion for image-to-image generation) require more time, taking approximately 3 days on COCO. However, the augmentations are created in advance only once, so their generation has no effect on the actual cost of the adversarial training.

\clearpage

\section{Additional Results}
\label{abl:eval_other_results}

\subsection{Image-text retrieval (ITR)}

In this section, we present comprehensive results for image-text retrieval (ITR), evaluating defense methods against unimodal attacks (PGD~\cite{madry2017towards} for image attack and BERT-attack~\cite{li2020bert}) as well as against multimodal attacks (Co-Attack~\cite{zhang2022towards} and SGA~\cite{lu2023set}).
Table~\ref{tab:app-clip-flickr-itr-all} and Table~\ref{tab:app-clip-coco-itr-all} present the results of CLIP trained on Flickr30k and COCO, respectively. Table~\ref{tab:app-albef-flickr-itr-all} and Table~\ref{tab:app-albef-coco-itr-all} show the results of ALBEF trained on Flickr30k and COCO, while Table~\ref{tab:app-blip-coco-itr-all} presents the results of BLIP trained on COCO.

First, the results demonstrate that \Mat{} consistently outperforms baseline unimodal AT methods in terms of multimodal robustness, which is crucial given that multimodal attacks are significantly stronger than unimodal attacks in VL tasks. The robustness against image attacks (PGD) is generally similar between image-only defenses (FARE or TeCoA-ITR) and \Mat{}, which aligns with expectations.

Second, the results indicate that effective augmentations, such as I2T(Human) for text augmentation and T-I2I(SD) for image augmentation, consistently improve multimodal robustness.
The important factors that determines the effectiveness of augmentations are described in Sec.~\ref{sec:results}: both I2T(Human) and T-I2I(SD) augment well-aligned image-text pairs with sufficient diversity, while keeping their distribution close to the original samples.

In addition to the main paper, we present results on many-to-many augmentation in Tab.~\ref{tab:app-clip-coco-itr-all}, where we combine image and text augmentations---specifically, T-I2I(SD) for image augmentation and I2T(Human) for text augmentation—--aiming at enhancing robustness.
However, our results show that this combination does not provide further improvements over using I2T(Human) alone, suggesting that the added image augmentation does not contribute to additional robustness.
A possible reason for this is that T-I2I(SD) introduces a slight distribution shift, which may offset its benefits.
These findings indicate that while many-to-many augmentations have theoretical potential, they require augmentation techniques that introduce sufficient diversity while maintaining consistency with the original data distribution.

\begin{table*}[ht]
\vspace{-2pt}
    \centering
    \small
    \resizebox{1.0\textwidth}{!}{%
\begin{tabular}{lll:ll:llll:llll}
\toprule
\multirow{3}{*}{\textbf{Method}} & \multirow{3}{*}{\textbf{Img aug.}} & \multirow{3}{*}{\textbf{Text. aug.}} &  & & \multicolumn{4}{c}{\textbf{Unimodal attacks}}  & \multicolumn{4}{c}{\textbf{Multimodal attacks}}  \\
 &  & & \multicolumn{2}{c}{\textbf{Clean}} & \multicolumn{2}{c}{\textbf{PGD (image attack)}}  & \multicolumn{2}{c}{\textbf{BERT (text attack)}} & \multicolumn{2}{c}{\textbf{Co-Attack}} & \multicolumn{2}{c}{\textbf{SGA}}  \\ \cmidrule(lr){4-5} \cmidrule(lr){6-7} \cmidrule(lr){8-9}  \cmidrule(lr){10-11}  \cmidrule(lr){12-13} 
& & & TR@1 & IR@1 & TR@1 & IR@1 & TR@1 & IR@1 & TR@1 & IR@1 & TR@1 & IR@1  \\ 
\midrule
\rowcolorbase Finetune &  &  & \textbf{92.1} & \textbf{77.2} & 11.9 & 10.1 & 75.4 & 53.1 & 11.0 & 6.7 & 0.6 & 0.6 \\ \hline
\rowcolorbase FARE &  &  & 75.9 & 61.0 & 69.7 & 55.1 & 53.2 & 40.2 & 41.8 & 30.7 & 27.1 & 21.0 \\
\rowcolorbase TeCoA-ITR &  &  & 83.1 & 68.2 & 77.7 & 61.9 & 64.7 & 42.7 & 52.9 & 31.9 & 27.5 & 17.6 \\ \hline
(ours) \Mat &  &  & 83.7 & 67.5 & 77.4 & 61.4 & 72.2 & 51.1 & 56.9 & 37.1 & 37.5 & 24.8 \\ \hdashline
\multirow{7}{*}{(ours) \AugMatAbbrev} &  & Basic(EDA) & 85.4 \arrowup{1.7} & 69.5 \arrowup{2.0} & 79.5 \arrowup{2.1} & 62.4 \arrowup{1.0} & 75.7 \arrowup{3.5} & 53.7 \arrowup{2.6} & 60.1 \arrowup{3.2} & 39.5 \arrowup{2.4} & 39.1 \arrowup{1.6} & 27.5 \arrowup{2.7} \\
 &  & T2T(LangRW) & 80.9 \arrowdown{2.8} & 67.0 \arrowdown{0.5} & 74.2 \arrowdown{3.2} & 60.1 \arrowdown{1.3} & 72.6 \arrowup{0.4} & 51.3 \arrowup{0.3} & 58.9 \arrowup{2.0} & 38.9 \arrowup{1.9} & 40.0 \arrowup{2.5} & 27.4 \arrowup{2.6} \\
 &  & I2T(div-Caps) & 84.7 \arrowup{1.0} & 69.2 \arrowup{1.7} & 79.0 \arrowup{1.6} & 64.5 \arrowup{3.1} & 74.8 \arrowup{2.6} & 51.4 \arrowup{0.3} & 60.3 \arrowup{3.4} & 39.6 \arrowup{2.6} & 40.3 \arrowup{2.8} & 27.8 \arrowup{3.0} \\
 &  & I2T(Human) & 85.7 \arrowup{2.0} & \underline{71.9} \arrowup{4.4} & \underline{82.0} \arrowup{4.6} & 65.9 \arrowup{4.5} & \textbf{77.6} \arrowup{5.4} & \textbf{56.3} \arrowup{5.2} & \textbf{63.3} \arrowup{6.4} & \textbf{44.0} \arrowup{6.9} & \textbf{45.6} \arrowup{8.1} & \textbf{32.2} \arrowup{7.4} \\
 & Basic(RandAug) &  & 84.1 \arrowup{0.4} & 67.1 \arrowdown{0.4} & 78.7 \arrowup{1.3} & 61.6 \arrowup{0.2} & 70.9 \arrowdown{1.3} & 50.4 \arrowdown{0.7} & 57.8 \arrowup{0.9} & 37.3 \arrowup{0.2} & 35.6 \arrowdown{1.9} & 24.4 \arrowdown{0.4} \\
 & T-I2I(SD) &  & 83.8 \arrowup{0.1} & 69.1 \arrowup{1.6} & 77.7 \arrowup{0.3} & 62.2 \arrowup{0.8} & 75.3 \arrowup{3.1} & 52.1 \arrowup{1.0} & 59.7 \arrowup{2.8} & 38.0 \arrowup{1.0} & 39.3 \arrowup{1.8} & 25.8 \arrowup{1.0} \\
 & T2I(SD) &  & 83.3 \arrowdown{0.4} & 68.4 \arrowup{0.9} & 76.1 \arrowdown{1.3} & 61.5 \arrowup{0.1} & 73.3 \arrowup{1.1} & 52.2 \arrowup{1.1} & 57.2 \arrowup{0.3} & 37.8 \arrowup{0.7} & 37.9 \arrowup{0.4} & 25.2 \arrowup{0.4} \\
\bottomrule
\end{tabular}
    }
    \vspace{-5pt}
    \caption{Comparison of CLIP trained on the Flickr30k dataset for ITR under the no-attack scenario (Clean), unimodal attacks (PGD and BERT), and multimodal attacks (Co-Attack and SGA), reporting R@k for text retrieval (TR@k) and image retrieval (IR@k).
    }
    \label{tab:app-clip-flickr-itr-all}
\end{table*}

\begin{table*}[ht]
\vspace{-2pt}
    \centering
    \small
    \resizebox{1.0\textwidth}{!}{%
\begin{tabular}{lllllllllllll}
\toprule
\multirow{3}{*}{\textbf{Method}} & \multirow{3}{*}{\textbf{Img aug.}} & \multirow{3}{*}{\textbf{Text. aug.}} &  & & \multicolumn{4}{c}{\textbf{Unimodal attacks}}  & \multicolumn{4}{c}{\textbf{Multimodal attacks}}  \\
 &  & & \multicolumn{2}{c}{\textbf{Clean}} & \multicolumn{2}{c}{\textbf{PGD (image attack)}}  & \multicolumn{2}{c}{\textbf{BERT (text attack)}} & \multicolumn{2}{c}{\textbf{Co-Attack}} & \multicolumn{2}{c}{\textbf{SGA}}  \\ \cmidrule(lr){4-5} \cmidrule(lr){6-7} \cmidrule(lr){8-9}  \cmidrule(lr){10-11}  \cmidrule(lr){12-13} 
& & & TR@1 & IR@1 & TR@1 & IR@1 & TR@1 & IR@1 & TR@1 & IR@1 & TR@1 & IR@1  \\ 
\midrule
\rowcolorbase Finetune &  &  & \textbf{66.6} & \textbf{50.1} & 6.8 & 5.4 & 36.9 & 23.7 & 2.9 & 1.8 & 0.1 & 0.1 \\ \hline
\rowcolorbase FARE &  &  & 45.2 & 32.3 & 40.4 & 29.3 & 22.4 & 15.9 & 16.7 & 11.4 & 9.1 & 6.9 \\
\rowcolorbase TeCoA-ITR &  &  & 58.0 & 41.6 & 51.0 & \underline{36.7} & 30.6 & 18.2 & 21.3 & 12.5 & 9.6 & 6.2 \\ \hline
(ours) \Mat &  &  & 55.8 & 40.7 & 49.5 & 36.1 & \underline{42.9} & 27.4 & 31.4 & 19.5 & 17.7 & 12.3 \\ \hdashline
\multirow{8}{*}{(ours) \AugMatAbbrev} &  & Basic(EDA) & 55.9 \arrowup{0.2} & 40.2 \arrowdown{0.5} & 48.6 \arrowdown{0.9} & 35.3 \arrowdown{0.8} & 41.1 \arrowdown{1.8} & 27.1 \arrowdown{0.4} & 29.8 \arrowdown{1.6} & 19.5 \arrowdown{0.0} & 18.4 \arrowup{0.7} & \underline{12.9} \arrowup{0.6} \\
 &  & T2T(LangRW) & 51.8 \arrowdown{4.0} & 37.7 \arrowdown{3.0} & 45.7 \arrowdown{3.8} & 33.5 \arrowdown{2.5} & 38.7 \arrowdown{4.2} & 26.1 \arrowdown{1.3} & 28.7 \arrowdown{2.7} & 19.0 \arrowdown{0.6} & 17.0 \arrowdown{0.7} & 12.6 \arrowup{0.3} \\
 &  & I2T(div-Caps) & 56.7 \arrowup{0.9} & 39.9 \arrowdown{0.8} & \underline{51.2} \arrowup{1.8} & 36.3 \arrowup{0.2} & 40.2 \arrowdown{2.6} & 26.1 \arrowdown{1.3} & 30.3 \arrowdown{1.1} & 19.4 \arrowdown{0.2} & \underline{18.9} \arrowup{1.2} & 12.5 \arrowup{0.2} \\
 &  & I2T(Human) & \underline{58.9} \arrowup{3.1} & \underline{43.1} \arrowup{2.4} & \textbf{52.2} \arrowup{2.7} & \textbf{38.5} \arrowup{2.4} & \textbf{44.7} \arrowup{1.9} & \textbf{29.9} \arrowup{2.4} & \textbf{33.6} \arrowup{2.2} & \textbf{21.9} \arrowup{2.3} & \textbf{21.3} \arrowup{3.6} & \textbf{14.5} \arrowup{2.2} \\
 & Basic(RandAug) &  & 55.9 \arrowup{0.1} & 40.7 \arrowdown{0.0} & 49.4 \arrowdown{0.1} & 36.3 \arrowup{0.2} & 41.7 \arrowdown{1.2} & 27.2 \arrowdown{0.2} & 30.7 \arrowdown{0.7} & 19.6 \arrowup{0.1} & 18.3 \arrowup{0.6} & 12.6 \arrowup{0.3} \\
 & T2I(SD) &  & 54.2 \arrowdown{1.6} & 38.3 \arrowdown{2.4} & 46.3 \arrowdown{3.2} & 33.4 \arrowdown{2.7} & 39.5 \arrowdown{3.3} & 25.9 \arrowdown{1.5} & 28.5 \arrowdown{2.9} & 18.1 \arrowdown{1.5} & 15.9 \arrowdown{1.8} & 11.0 \arrowdown{1.3} \\
 & T,I2I(SD) &  & 57.1 \arrowup{1.3} & 41.7 \arrowup{1.0} & 49.7 \arrowup{0.2} & 36.0 \arrowdown{0.1} & 41.9 \arrowdown{0.9} & 28.3 \arrowup{0.8} & 31.3 \arrowdown{0.0} & \underline{20.2} \arrowup{0.7} & 18.8 \arrowup{1.1} & 12.6 \arrowup{0.2} \\
 \quad \: \textit{(N:N)} & T,I2I(SD) & I2T(Human) & \underline{59.0} \arrowup{3.2} & \underline{43.3} \arrowup{2.6} & \underline{52.2} \arrowup{2.7} & \underline{38.0} \arrowup{1.9} & \underline{44.6} \arrowup{1.7} & \underline{29.8} \arrowup{2.3} & \textbf{34.1} \arrowup{2.8} & \underline{21.2} \arrowup{1.7} & \underline{20.8} \arrowup{3.1} & \underline{14.2} \arrowup{1.9} \\
\bottomrule
\end{tabular}
    }
    \vspace{-5pt}
    \caption{Comparison of CLIP trained on the COCO dataset for ITR under the no-attack scenario (Clean), unimodal attacks (PGD and BERT), and multimodal attacks (Co-Attack and SGA), reporting R@k for text retrieval (TR@k) and image retrieval (IR@k).
    }
    \label{tab:app-clip-coco-itr-all}
\end{table*}

\begin{table*}[ht]
\vspace{-2pt}
    \centering
    \small
    \resizebox{1.0\textwidth}{!}{%
\begin{tabular}{lllllllllllll}
\toprule
\multirow{3}{*}{\textbf{Method}} & \multirow{3}{*}{\textbf{Img aug.}} & \multirow{3}{*}{\textbf{Text. aug.}} &  & & \multicolumn{4}{c}{\textbf{Unimodal attacks}}  & \multicolumn{4}{c}{\textbf{Multimodal attacks}}  \\
 &  & & \multicolumn{2}{c}{\textbf{Clean}} & \multicolumn{2}{c}{\textbf{PGD (image attack)}}  & \multicolumn{2}{c}{\textbf{BERT (text attack)}} & \multicolumn{2}{c}{\textbf{Co-Attack}} & \multicolumn{2}{c}{\textbf{SGA}}  \\ \cmidrule(lr){4-5} \cmidrule(lr){6-7} \cmidrule(lr){8-9}  \cmidrule(lr){10-11}  \cmidrule(lr){12-13} 
& & & TR@1 & IR@1 & TR@1 & IR@1 & TR@1 & IR@1 & TR@1 & IR@1 & TR@1 & IR@1  \\ 
\midrule
\rowcolorbase Finetune &  &  & \textbf{89.5} & \textbf{77.7} & 48.8 & 34.0 & 79.1 & 55.0 & 32.3 & 18.4 & 2.5 & 1.3 \\ \hline
\rowcolorbase TeCoA-ITR &  &  & 85.4 & 69.3 & 78.7 & 63.1 & 72.1 & 48.9 & 61.0 & 38.1 & 35.5 & 21.9 \\ \hline
(ours) \Mat &  &  & 82.0 & 66.3 & 78.0 & 63.8 & 77.1 & 56.1 & 73.0 & 54.0 & 47.1 & 32.9 \\ \hdashline
\multirow{7}{*}{(ours) \AugMatAbbrev} &  & Basic(EDA) & 82.2 \arrowup{0.2} & 67.9 \arrowup{1.6} & 78.2 \arrowup{0.2} & 63.8 \arrowup{0.0} & 78.6 \arrowup{1.5} & 58.1 \arrowup{2.0} & 71.1 \arrowdown{1.9} & 52.9 \arrowdown{1.1} & 44.6 \arrowdown{2.5} & 31.2 \arrowdown{1.7} \\
 &  & T2T(LangRW) & 81.4 \arrowdown{0.6} & 67.5 \arrowup{1.3} & 76.0 \arrowdown{2.0} & 62.5 \arrowdown{1.3} & 75.8 \arrowdown{1.3} & 56.2 \arrowup{0.1} & 70.0 \arrowdown{3.0} & 51.1 \arrowdown{3.0} & 46.3 \arrowdown{0.8} & 32.4 \arrowdown{0.5} \\
 &  & I2T(div-Caps) & 85.6 \arrowup{3.6} & 71.0 \arrowup{4.8} & \textbf{81.7} \arrowup{3.7} & 66.6 \arrowup{2.8} & \underline{79.7} \arrowup{2.6} & \underline{60.1} \arrowup{3.9} & \underline{75.6} \arrowup{2.6} & 55.2 \arrowup{1.2} & 48.8 \arrowup{1.7} & 35.0 \arrowup{2.1} \\
 &  & I2T(Human) & \underline{85.8} \arrowup{3.8} & \underline{72.8} \arrowup{6.5} & 80.2 \arrowup{2.2} & \textbf{67.6} \arrowup{3.8} & \textbf{81.5} \arrowup{4.4} & \textbf{62.8} \arrowup{6.7} & \textbf{77.5} \arrowup{4.5} & \textbf{58.7} \arrowup{4.7} & \underline{52.9} \arrowup{5.8} & \textbf{38.8} \arrowup{5.9} \\
 & Basic(RandAug) &  & 82.2 \arrowup{0.2} & 67.2 \arrowup{0.9} & 78.3 \arrowup{0.3} & 63.2 \arrowdown{0.6} & 76.9 \arrowdown{0.2} & 56.2 \arrowup{0.1} & 72.7 \arrowdown{0.3} & 53.2 \arrowdown{0.8} & 48.3 \arrowup{1.2} & 33.4 \arrowup{0.5} \\
 & T-I2I(SD) &  & 85.1 \arrowup{3.1} & 69.8 \arrowup{3.6} & 81.7 \arrowup{3.7} & \underline{67.1} \arrowup{3.3} & 79.4 \arrowup{2.3} & 59.1 \arrowup{2.9} & 74.9 \arrowup{1.9} & \underline{55.9} \arrowup{1.8} & \textbf{55.2} \arrowup{8.1} & \underline{37.6} \arrowup{4.8} \\
 & T2I(SD) &  & 83.7 \arrowup{1.7} & 68.3 \arrowup{2.1} & 79.5 \arrowup{1.5} & 63.6 \arrowdown{0.2} & 76.7 \arrowdown{0.4} & 59.0 \arrowup{2.9} & 72.4 \arrowdown{0.6} & 55.1 \arrowup{1.1} & 52.0 \arrowup{4.9} & 36.2 \arrowup{3.3} \\
\bottomrule
\end{tabular}
    }
    \vspace{-5pt}
    \caption{Comparison of ALBEF trained on the Flickr30k dataset for ITR under the no-attack scenario (Clean), unimodal attacks (PGD and BERT), and multimodal attacks (Co-Attack and SGA), reporting R@k for text retrieval (TR@k) and image retrieval (IR@k).
    }
    \label{tab:app-albef-flickr-itr-all}
\end{table*}

\begin{table*}[ht]
\vspace{-2pt}
    \centering
    \small
    \resizebox{1.0\textwidth}{!}{%
\begin{tabular}{lllllllllllll}
\toprule
\multirow{3}{*}{\textbf{Method}} & \multirow{3}{*}{\textbf{Img aug.}} & \multirow{3}{*}{\textbf{Text. aug.}} &  & & \multicolumn{4}{c}{\textbf{Unimodal attacks}}  & \multicolumn{4}{c}{\textbf{Multimodal attacks}}  \\
 &  & & \multicolumn{2}{c}{\textbf{Clean}} & \multicolumn{2}{c}{\textbf{PGD (image attack)}}  & \multicolumn{2}{c}{\textbf{BERT (text attack)}} & \multicolumn{2}{c}{\textbf{Co-Attack}} & \multicolumn{2}{c}{\textbf{SGA}}  \\ \cmidrule(lr){4-5} \cmidrule(lr){6-7} \cmidrule(lr){8-9}  \cmidrule(lr){10-11}  \cmidrule(lr){12-13} 
& & & TR@1 & IR@1 & TR@1 & IR@1 & TR@1 & IR@1 & TR@1 & IR@1 & TR@1 & IR@1  \\ 
\midrule
\rowcolorbase Finetune &  &  & \textbf{69.9} & \textbf{53.6} & 30.0 & 17.4 & 49.0 & 31.4 & 17.2 & 9.1 & 1.0 & 0.7 \\ \hline
\rowcolorbase TeCoA-ITR &  &  & 64.8 & 48.6 & 49.5 & 38.1 & 48.5 & 28.5 & 38.2 & 19.4 & 14.2 & 9.5 \\ \hline
(ours) \Mat &  &  & 63.9 & 46.2 & 60.2 & 38.3 & 55.4 & 36.9 & 52.9 & 32.1 & 31.2 & 21.2 \\ \hdashline
\multirow{7}{*}{(ours) \AugMatAbbrev} &  & Basic(EDA) & 63.9 \arrowup{0.0} & 46.8 \arrowup{0.6} & 53.2 \arrowdown{7.0} & 38.7 \arrowup{0.4} & 54.4 \arrowdown{1.0} & 36.4 \arrowdown{0.5} & 50.8 \arrowdown{2.2} & 30.9 \arrowdown{1.2} & 31.5 \arrowup{0.3} & 20.9 \arrowdown{0.3} \\
 &  & T2T(LangRW) & 59.7 \arrowdown{4.2} & 42.1 \arrowdown{4.1} & 46.6 \arrowdown{13.5} & 31.0 \arrowdown{7.3} & 51.4 \arrowdown{4.1} & 32.9 \arrowdown{4.0} & 34.6 \arrowdown{18.3} & 23.9 \arrowdown{8.2} & 25.4 \arrowdown{5.8} & 15.7 \arrowdown{5.5} \\
 &  & I2T(div-Caps) & 66.0 \arrowup{2.0} & \underline{49.9} \arrowup{3.7} & 57.0 \arrowdown{3.2} & 37.9 \arrowdown{0.4} & \underline{56.6} \arrowup{1.2} & 38.5 \arrowup{1.6} & 47.5 \arrowdown{5.4} & 26.4 \arrowdown{5.6} & \underline{35.5} \arrowup{4.3} & 20.3 \arrowdown{0.9} \\
 &  & I2T(Human) & \underline{68.5} \arrowup{4.5} & 49.1 \arrowup{3.0} & \textbf{64.9} \arrowup{4.7} & 37.3 \arrowdown{1.0} & \textbf{59.7} \arrowup{4.3} & \textbf{39.9} \arrowup{3.0} & \textbf{55.3} \arrowup{2.4} & 29.3 \arrowdown{2.8} & \textbf{36.2} \arrowup{4.9} & \textbf{23.5} \arrowup{2.2} \\
 & Basic(RandAug) &  & 63.1 \arrowdown{0.9} & 48.3 \arrowup{2.1} & 61.1 \arrowup{0.9} & \textbf{46.0} \arrowup{7.7} & 54.8 \arrowdown{0.7} & 38.3 \arrowup{1.4} & 52.1 \arrowdown{0.8} & \underline{35.7} \arrowup{3.7} & 30.3 \arrowdown{1.0} & 21.7 \arrowup{0.4} \\
 & T2I(SD) &  & 61.5 \arrowdown{2.4} & 46.0 \arrowdown{0.1} & 58.7 \arrowdown{1.5} & 37.8 \arrowdown{0.5} & 53.5 \arrowdown{2.0} & 37.5 \arrowup{0.6} & 50.6 \arrowdown{2.3} & 33.0 \arrowup{1.0} & 25.3 \arrowdown{5.9} & 18.6 \arrowdown{2.6} \\
 & T,I2I(SD) &  & 64.9 \arrowup{0.9} & 47.1 \arrowup{1.0} & \underline{61.8} \arrowup{1.6} & 36.1 \arrowdown{2.2} & 56.5 \arrowup{1.1} & 38.1 \arrowup{1.3} & 54.1 \arrowup{1.2} & 29.2 \arrowdown{2.9} & 33.2 \arrowup{2.0} & \underline{22.4} \arrowup{1.2} \\
\bottomrule
\end{tabular}
    }
    \vspace{-5pt}
    \caption{Comparison of ALBEF trained on the COCO dataset for ITR under the no-attack scenario (Clean), unimodal attacks (PGD and BERT), and multimodal attacks (Co-Attack and SGA), reporting R@k for text retrieval (TR@k) and image retrieval (IR@k).
    }
    \label{tab:app-albef-coco-itr-all}
\end{table*}

\begin{table*}[ht]
\vspace{-2pt}
    \centering
    \small
    \resizebox{1.0\textwidth}{!}{%
\begin{tabular}{lllllllllllll}
\toprule
\multirow{3}{*}{\textbf{Method}} & \multirow{3}{*}{\textbf{Img aug.}} & \multirow{3}{*}{\textbf{Text. aug.}} &  & & \multicolumn{4}{c}{\textbf{Unimodal attacks}}  & \multicolumn{4}{c}{\textbf{Multimodal attacks}}  \\
 &  & & \multicolumn{2}{c}{\textbf{Clean}} & \multicolumn{2}{c}{\textbf{PGD (image attack)}}  & \multicolumn{2}{c}{\textbf{BERT (text attack)}} & \multicolumn{2}{c}{\textbf{Co-Attack}} & \multicolumn{2}{c}{\textbf{SGA}}  \\ \cmidrule(lr){4-5} \cmidrule(lr){6-7} \cmidrule(lr){8-9}  \cmidrule(lr){10-11}  \cmidrule(lr){12-13} 
& & & TR@1 & IR@1 & TR@1 & IR@1 & TR@1 & IR@1 & TR@1 & IR@1 & TR@1 & IR@1  \\ 
\midrule
\rowcolorbase Finetune &  &  & \textbf{72.9} & \textbf{57.5} & 50.9 & 35.7 & 39.2 & 27.5 & 18.4 & 9.9 & 1.2 & 1.1 \\ \hline
\rowcolorbase TeCoA-ITR &  &  & 64.6 & 51.8 & 63.8& 50.3 & 37.6 & 24.9 & 31.2& 20.2 & 20.2 & 13.9 \\ \hline
(ours) \Mat &  &  & 66.9 & 49.9 & \underline{67.8} & 49.5 & 54.7 & \underline{37.1} & \underline{50.5} & \underline{32.8} & 31.3 & 21.0 \\ \hdashline
\multirow{2}{*}{(ours) \AugMatAbbrev} &  & I2T(Human) & \underline{71.0} \arrowup{4.1} & \underline{54.3} \arrowup{4.5} & \textbf{68.6} \arrowup{0.8} & \textbf{52.1} \arrowup{2.6} & \textbf{57.1} \arrowup{2.4} & \textbf{40.4} \arrowup{3.3} & \textbf{51.2} \arrowup{0.7} & \textbf{35.3} \arrowup{2.5} & \textbf{35.6} \arrowup{4.3} & \textbf{25.7} \arrowup{4.7} \\
 & T,I2I(SD) &  & 68.2 \arrowup{1.3} & 50.5 \arrowup{0.6} & 62.3 \arrowdown{5.5} & 43.6 \arrowdown{6.0} & \underline{55.0} \arrowup{0.2} & 37.0 \arrowdown{0.0} & 47.9 \arrowdown{2.6} & 29.9 \arrowdown{3.0} & \underline{33.5} \arrowup{2.2} & \underline{22.9} \arrowup{1.9} \\
\bottomrule
\end{tabular}
    }
    \vspace{-5pt}
    \caption{Comparison of BLIP trained on the COCO dataset for ITR under the no-attack scenario (Clean), unimodal attacks (PGD and BERT), and multimodal attacks (Co-Attack and SGA), reporting R@k for text retrieval (TR@k) and image retrieval (IR@k).
    }
    \label{tab:app-blip-coco-itr-all}
\end{table*}

\clearpage

\subsection{Visual grounding (VG)}

Table~\ref{tab:app-albef-coco-vg-all} presents results of ALBEF trained on the COCO dataset for the VG task.
The results highlight \Mat's effectiveness beyond ITR task, as well as the effectiveness of one-to-many augmentations.
We observe that I2T(div-Caps) outperforms I2T(Human), which can be attributed to the design of the prompts used in I2T(div-Caps).
Specifically, I2T(div-Caps) generates captions that focus on both foreground and background objects, providing a more comprehensive description of the image (see Sec.~\ref{appendix:details_augmentation} for the prompt design). 
This makes I2T(div-Caps) particularly useful for the VG task, where understanding diverse objects located in different areas of the image is crucial.

\begin{table*}[ht]
\vspace{-2pt}
    \centering
    \small
\begin{tabular}{lll:ccc:ccc}
\toprule
\multirow{2}{*}{\textbf{Method}} & \multirow{2}{*}{\textbf{Img aug.}} & \multirow{2}{*}{\textbf{Text. aug.}} & \multicolumn{3}{c}{\textbf{Clean}}  & \multicolumn{3}{c}{\textbf{SGA}}  \\
\cmidrule(lr){4-6} \cmidrule(lr){7-9} 
 &  & & val & test-A  & test-B & val & test-A  & test-B \\ 
\midrule
\rowcolorbase Fine-tune &  &  & 50.2 & \textbf{57.6} & 40.6 & 32.9 & 36.3 & 29.6 \\
\rowcolorbase  TeCoA-ITR &  &  & 49.6 & 55.4 & 41.4 & 38.8 & 44.8 & 32.9 \\ \hline
(ours) \Mat &  &  & 48.8 & 53.3 & 40.7 & 40.4 & 44.2 & 35.1 \\ \hdashline
\multirow{3}{*}{(ours) \AugMatAbbrev} &  & I2T(div-Caps) & \textbf{51.0} \arrowup{2.2} & \underline{57.0} \arrowup{3.7} & \textbf{43.3} \arrowup{2.6} & \textbf{43.2} \arrowup{2.8} & \textbf{48.1} \arrowup{3.9} & \textbf{37.1} \arrowup{2.1} \\
 &  & I2T(Human) & \underline{50.6} \arrowup{1.8} & 55.9 \arrowup{2.6} & \underline{41.7} \arrowup{1.1} & \underline{42.0} \arrowup{1.6} & \underline{46.5} \arrowup{2.3} & \underline{35.7} \arrowup{0.6} \\
 & T,I2I(SD) &  & 49.6 \arrowup{0.7} & 55.9 \arrowup{2.6} & 40.8 \arrowup{0.1} & 41.1 \arrowup{0.7} & 46.0 \arrowup{1.8} & 34.8 \arrowdown{0.3} \\
\bottomrule
\end{tabular}
    \vspace{-5pt}
    \caption{Accuracy comparison of ALBEF trained on the COCO dataset for VG under the no-attack scenario (Clean) and multimodal attack (SGA).
    }
    \label{tab:app-albef-coco-vg-all}
\end{table*}

\clearpage

\subsection{Effectiveness of MAT+ vs.\ naively increasing data samples}
\label{appendix:fixed_data_size}

\begin{figure*}[ht]
    \centering
    \includegraphics[width=0.9\textwidth]{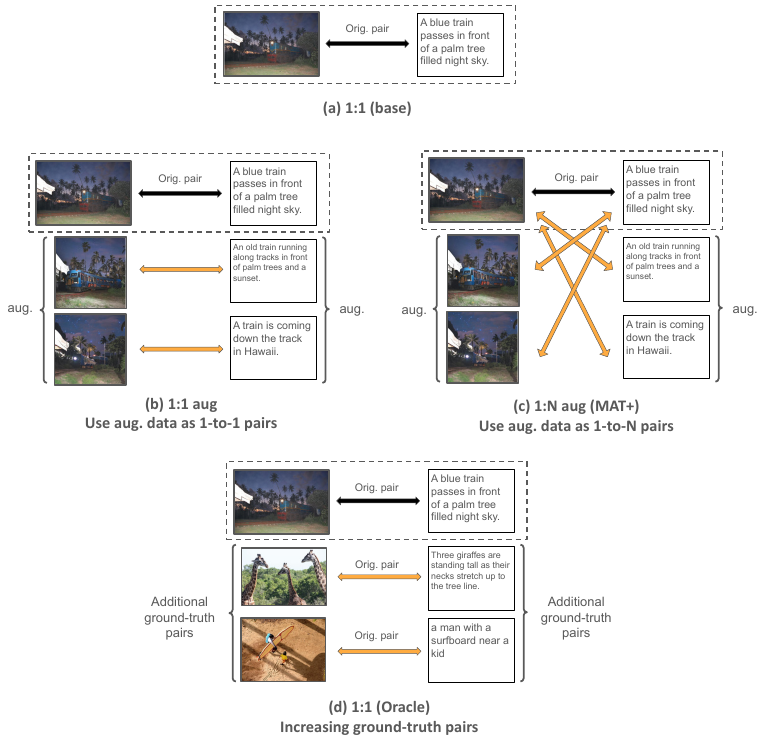}
    \caption{\textbf{Different ways of increasing the number of samples for adversarial training. (a) Original data samples (no increasing), (b) Augmenting orig. data samples without 1:N modeling, (c) Augmenting orig. data samples with 1:N modeling (MAT+), (d) Adding new orig. data samples (oracle).}}
    \label{fig:controlled_exp}
\end{figure*}

\subsubsection{Experimental Settings}
We conducted controlled experiments to disentangle the effect of data size from the one-to-many training strategy.  
Suppose we have $M$ image-text pairs and generate two augmented samples per modality, yielding $M$ original and $2M$ augmented images and texts (e.g., 30k original and 60k augmentations, in total of 90k).  
Then, we compare four settings (see Fig.~\ref{fig:controlled_exp}):
\begin{itemize}
    \item (a) \textbf{1:1 (base):} $M$ original ground-truth pairs.  
    \item (b) \textbf{1:1 aug:} Adding $2M$ augmentations and creating additional 1:1 pairs by directly combining augmented images and texts.
    \item (c) \textbf{1:N aug (MAT+):} Adding the same $2M$ augmentations as in (b), but each original sample forms 1-to-3 and 3-to-1 pairs with its two augmentations.
    \item (d) \textbf{1:1 (oracle):} Adding $2M$ extra original ground-truth pairs (no augmentations), serving as an upper bound.  
\end{itemize}
This setup enables a fair comparison between naively adding new augmentations (1:1) and using MAT+ (1:N) under the same data (i.e., (b) and (c)), while also contrasting them with an upper-bound oracle.

The base set ($M=30k$) was sampled randomly from COCO's $\sim$120k image-text training pairs. Then, we generated $60k$ augmentations using ``T-I2I(SD)'' for images and ``I2T(Human)'' for texts, which were the best performing techniques for each modality in our analysis. On the other hand, the oracle setting adds $60k$ randomly sampled ground-truth pairs, for a total of $90k$ original pairs.

\subsubsection{Results}

\begin{table}[ht]
\centering
\resizebox{1.0\textwidth}{!}{%
\begin{tabular}{lcccclllllllllll}
\toprule
 & \begin{tabular}[c]{@{}l@{}}Num.\\Orig.\\(GT)\end{tabular} & \begin{tabular}[c]{@{}l@{}}Num.\\Aug.\end{tabular} & \multicolumn{2}{c}{Augmentation} & \multicolumn{2}{c}{Clean} & \multicolumn{2}{c}{PGD} & \multicolumn{2}{c}{BERT-Attack} & \multicolumn{2}{c}{\textbf{SGA}} \\ \cmidrule(lr){4-5} \cmidrule(lr){6-7} \cmidrule(lr){8-9}  \cmidrule(lr){10-11}  \cmidrule(lr){12-13} 
 &  &  & Img. & Text & \multicolumn{1}{c}{IR@1} & \multicolumn{1}{c}{TR@1} & \multicolumn{1}{c}{IR@1} & \multicolumn{1}{c}{TR@1} & \multicolumn{1}{c}{IR@1} & \multicolumn{1}{c}{TR@1} & \multicolumn{1}{c}{IR@1} & \multicolumn{1}{c}{TR@1} \\ \midrule
(a) 1:1 (base) & 30k & - & - & - & 45.0 & 32.0 & 41.0 & 29.8 & 32.2 & 20.9 & 10.8 & 7.3 \\ \hline \hline
\rowcolorbase (b) 1:1 & 30k & 60k & T-I2I(SD) & I2T(Human) & 50.2 & 36.4 & 46.4 & 33.5 & 37.5 & 24.3 & 15.5 & 10.6 \\ \hdashline
\rowcolorbase (c) MAT+ (ours) & 30k & 60k & T-I2I(SD) & I2T(Human) & 54.6 \arrowup{4.4} & 39.0 \arrowup{2.6} & 49.3 \arrowup{2.9} & 36.1 \arrowup{2.6} & 40.1 \arrowup{2.6} & 26.3 \arrowup{2.0} & 16.3 \arrowup{0.8} & 11.4 \arrowup{0.8} \\ 
\multicolumn{15}{l}{\textit{Augmentation ablation}} \\
& 30k & 60k & T-I2I(SD) & \textit{EDA} & 49.5 & 35.5 & 45.3 & 32.6 & 36.5 & 23.2 & 13.2 & 9.2\\ 
& 30k & 60k & \textit{RandAug} & I2T(Human) & 52.7 & 38.0 & 48.8 & 35.2 & 38.4 & 24.9 & 15.3 & 10.6 \\ 
\hline \hline
(d) 1:1 (oracle) & 90k & - & - & - & 53.9 & 39.8 & 50.0 & 36.3 & 41.1 & 26.7 & 16.1 & 11.8 \\ 
\bottomrule
\end{tabular}
}
\caption{Controlled experiments to disentangle the effects of data size and one-to-many (1:N) augmentation (see Fig.~\ref{fig:controlled_exp} for augmentation types a$\sim$d). 
Compared to simply adding augmented pairs as 1:1 (b), using them as 1:N pairs (c, MAT+) yields larger robustness gains, nearly matching the oracle 1:1 setting (d).
}
\label{tab:controlled_exp}
\end{table}

Tab.~\ref{tab:controlled_exp} shows that while adding synthetic data as 1:1 pairs improves robustness, using them as 1:N pairs yields substantially larger gains, nearly matching the oracle 1-to-1 setting.  
This confirms that robustness improvements come not only from increased data size but also from \textbf{ambiguity modeling through 1-to-N alignment}.  

Besides ambiguity modeling, a key contribution of our work is identifying the properties of effective augmentations in adversarial defense---\textbf{high alignment}, \textbf{high diversity}, and \textbf{small distribution gap}.
In Tab.~\ref{tab:controlled_exp}(c) \textit{Augmentation ablation}, naive augmentations (RandAug and EDA) yield only limited improvements compared with higher-quality augmentations (T-I2I(SD) and I2T(Human)).

In summary, naively increasing the number of multimodal pairs is suboptimal compared with our proposed method, whose robustness is comparable to expanding the dataset by collecting new ground-truth samples. Also note that using MAT+ on (d) further boosts the accuracy, as shown in the main text when using the entire COCO dataset.

\color{black}

\clearpage

\section{Qualitative Results}
\begin{figure*}[ht]
    \centering
        \begin{subfigure}{1.0\columnwidth}
            \includegraphics[width=1.0\textwidth]{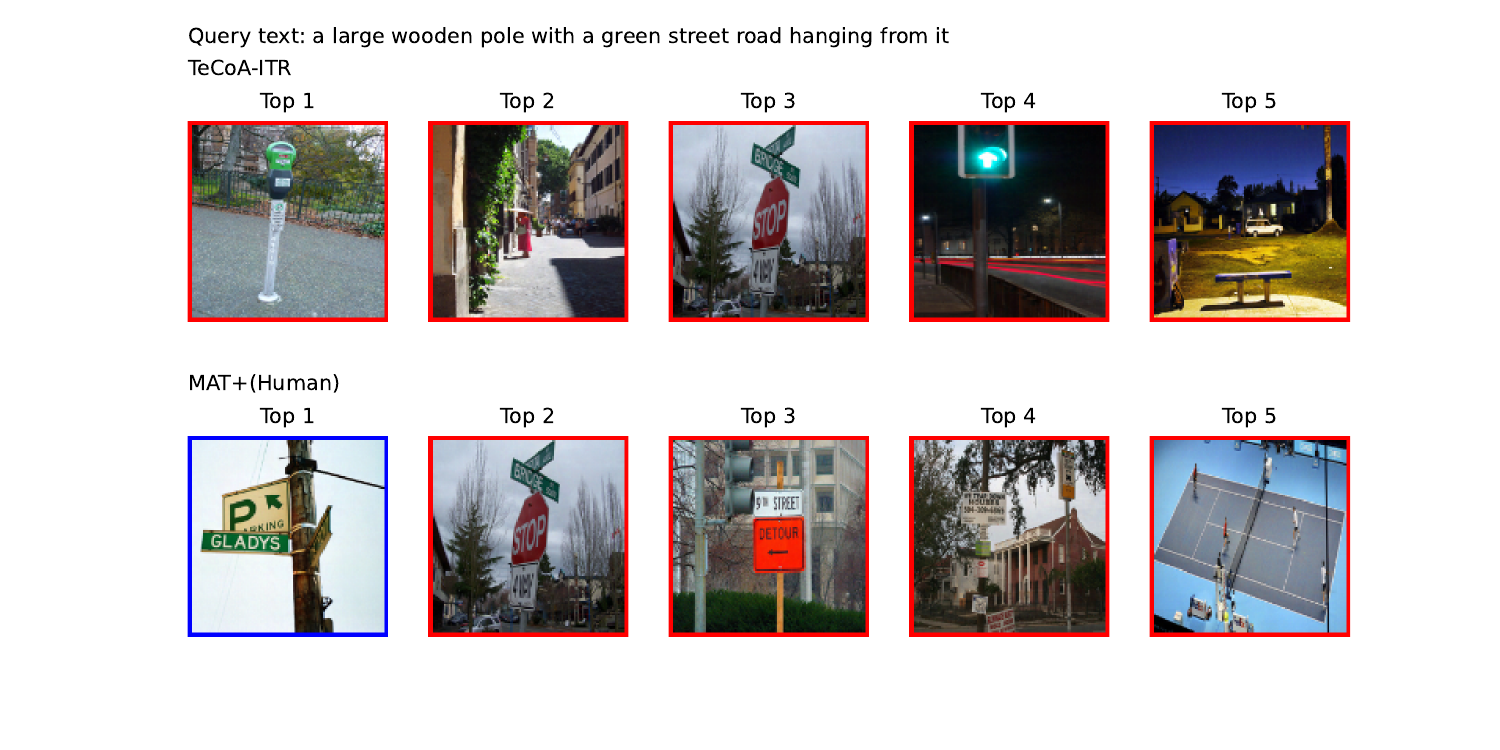}
        \end{subfigure}
        \begin{subfigure}{1.0\columnwidth}
            \includegraphics[width=1.0\textwidth]{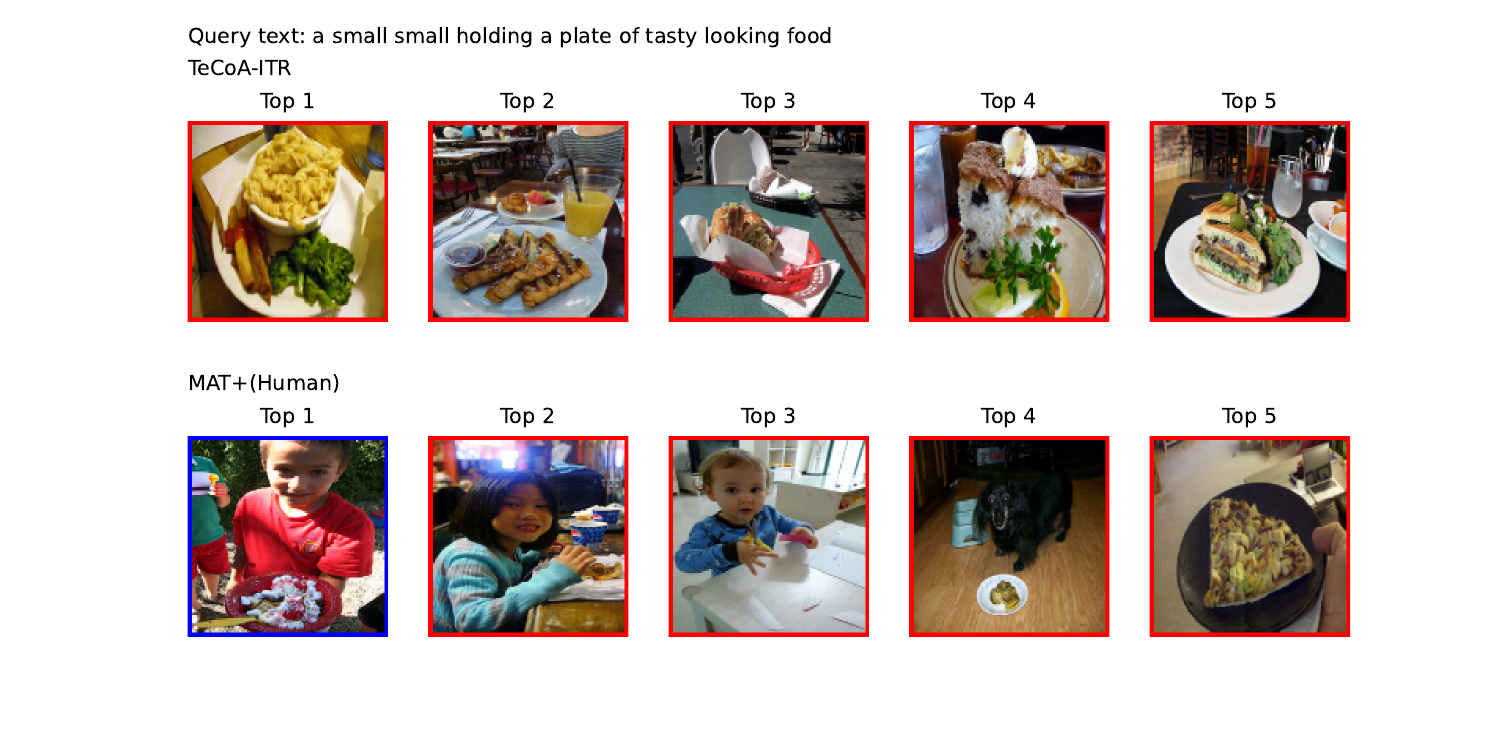}
        \end{subfigure}
    \caption{\textbf{Qualitative comparison of image retrieval} results under multimodal attack (SGA). We compare the TeCoA-ITR baseline (unimodal defense) with our proposed MAT+ (multimodal defense using I2T(Human) augmentations). Images with a blue border indicate correct retrieval, while those with a red border indicate incorrect retrieval.}
    \label{fig:vis_image_retrival}
\end{figure*}

\begin{figure*}[ht]
    \centering
        \begin{subfigure}{1.0\columnwidth}
            \includegraphics[width=1.0\textwidth]{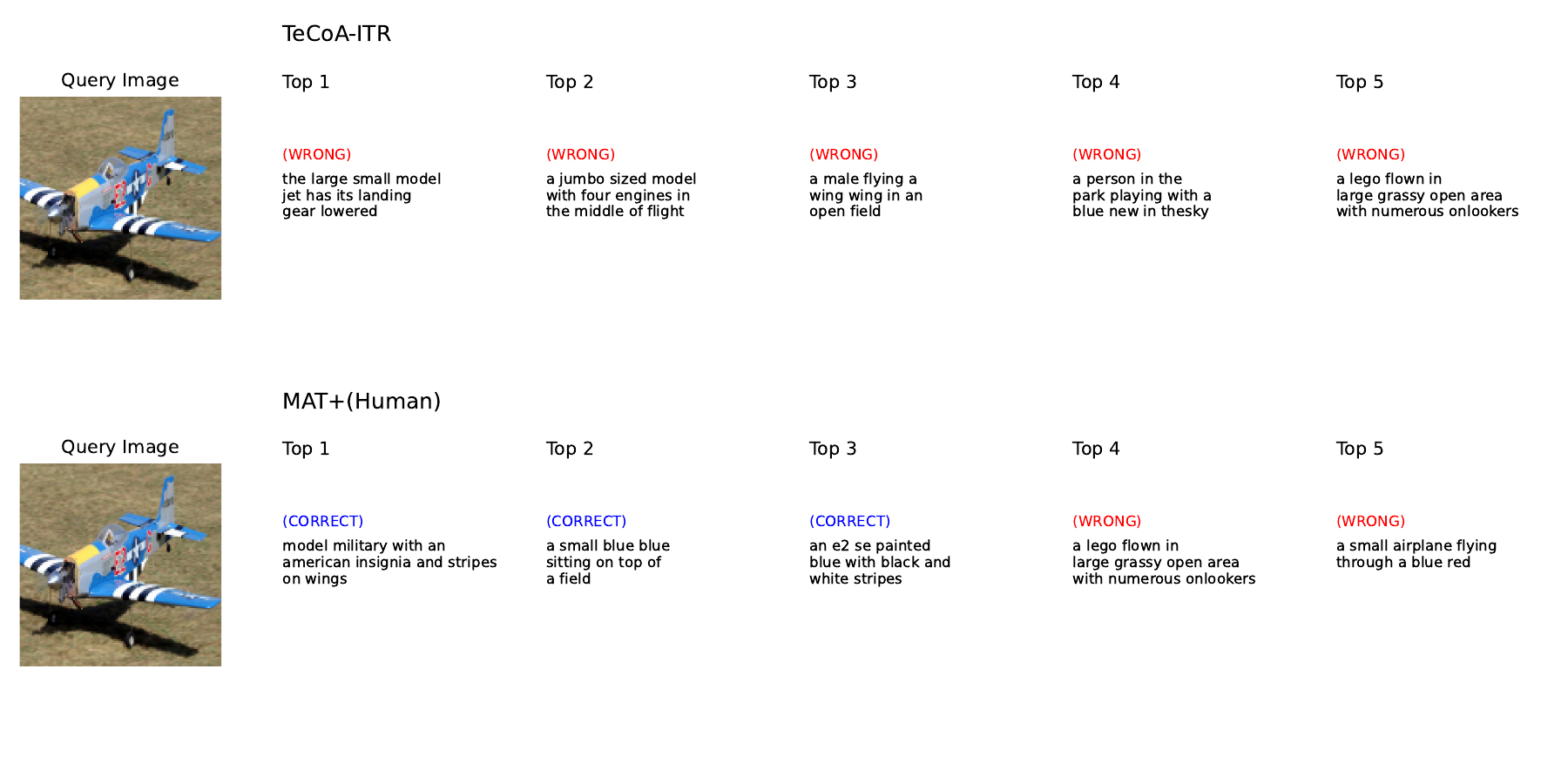}
        \end{subfigure}
        \begin{subfigure}{1.0\columnwidth}
            \includegraphics[width=1.0\textwidth]{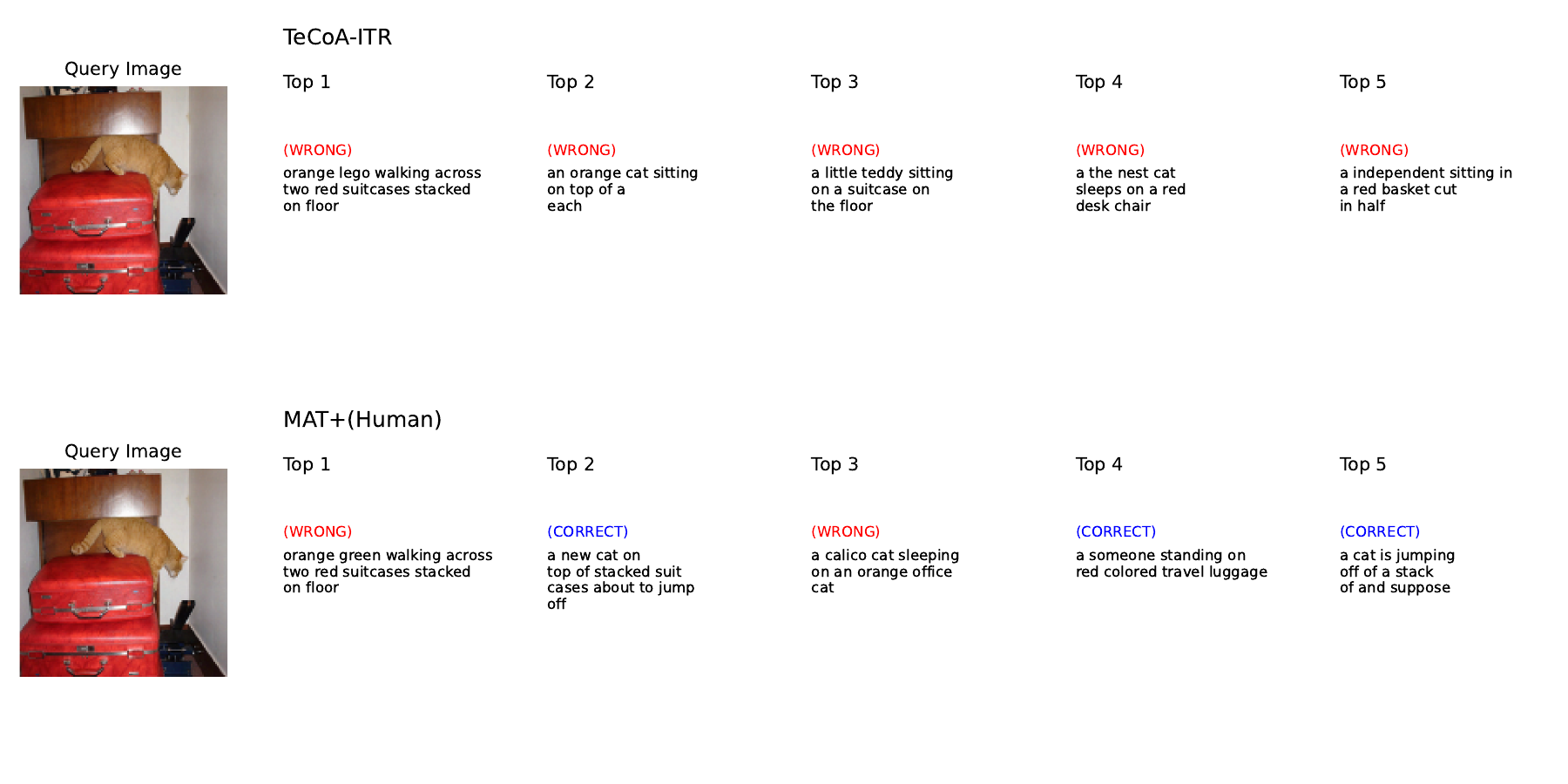}
        \end{subfigure}
    \caption{\textbf{Qualitative comparison of text retrieval} results under multimodal attack (SGA). We compare the TeCoA-ITR baseline (unimodal defense) with our proposed MAT+ (multimodal defense using I2T(Human) augmentations).}
    \label{fig:vis_text_retrival}
\end{figure*}

Here, we present qualitative results for ITR.
We compare the TeCoA-ITR baseline, a unimodal image defense method, with our proposed multimodal defense method, MAT+, which incorporates I2T(Human) augmentations. Both methods are evaluated against a multimodal attack (SGA).
Figure~\ref{fig:vis_image_retrival} and Figure~\ref{fig:vis_text_retrival} illustrate the comparison for image retrieval and text retrieval results, respectively.

\section{Visualization of Augmentations}
\label{sec:vis_aug}
\begin{figure*}[ht]
    \centering
    \includegraphics[width=1.0\textwidth]{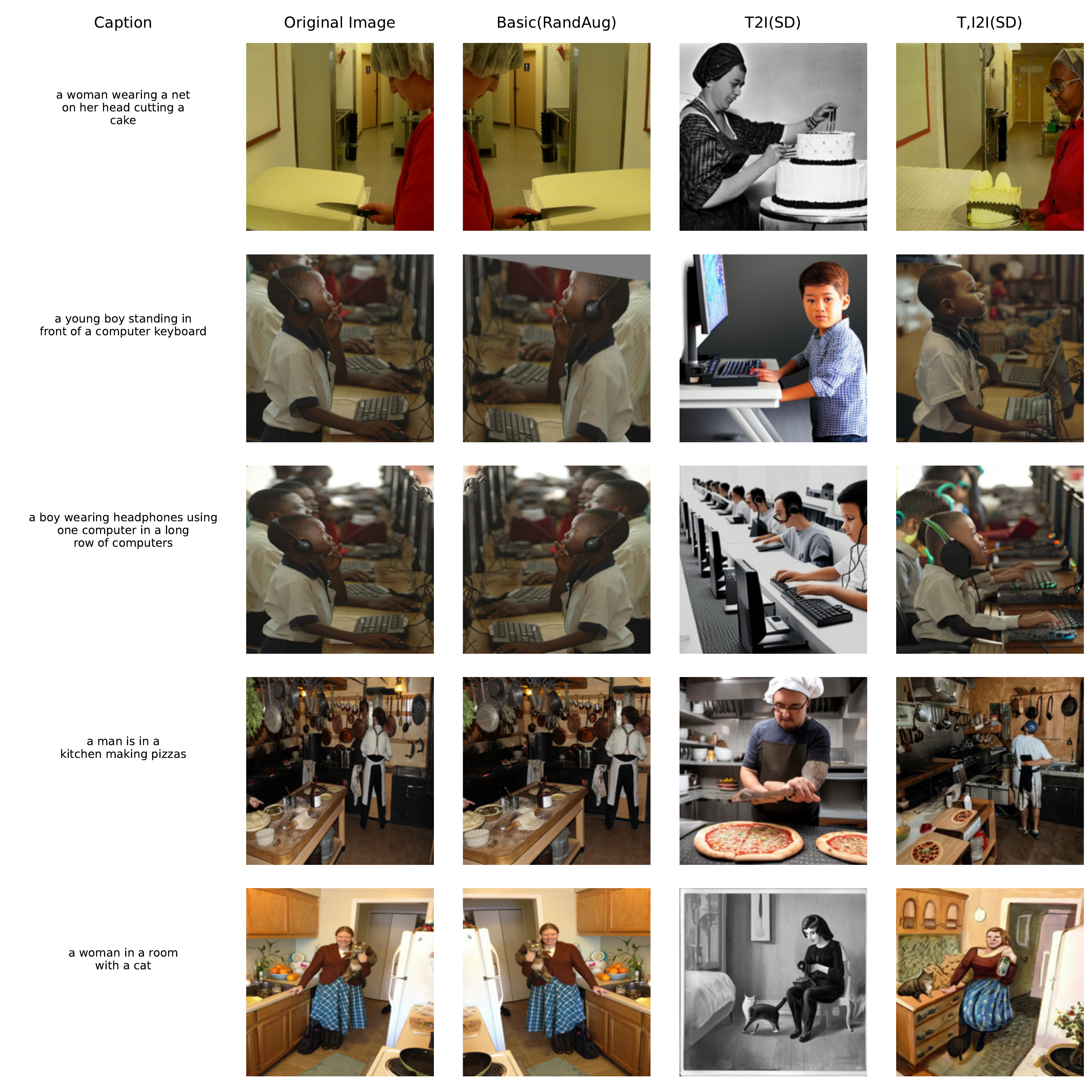}
    \vspace{-10pt}
    \caption{\textbf{Visualization of image augmentations (1).}}
    \label{fig:vis_aug_image_0}
\vspace{-5pt}
\end{figure*}

\begin{figure*}[ht]
    \centering
    \includegraphics[width=1.0\textwidth]{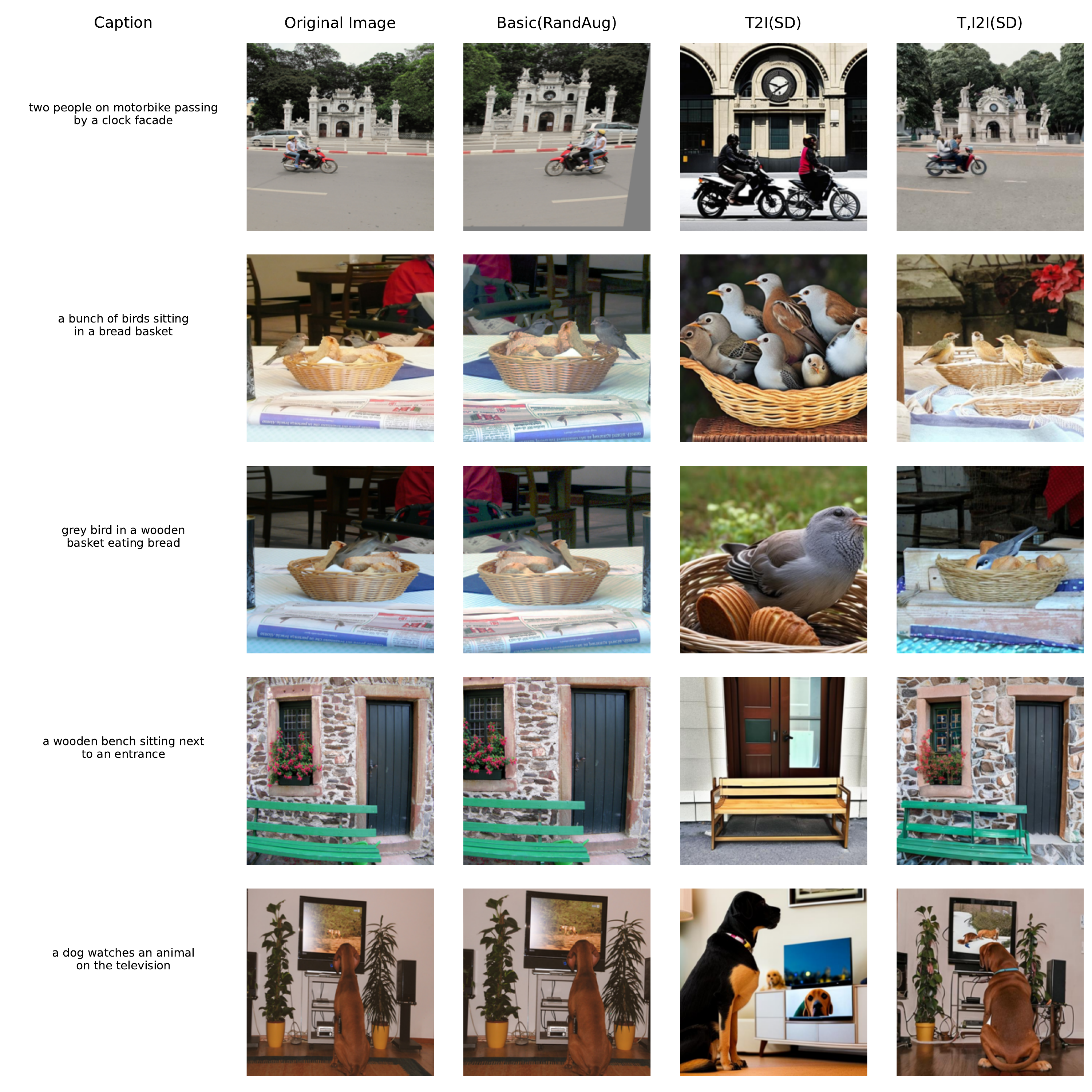}
    \vspace{-10pt}
    \caption{\textbf{Visualization of image augmentations (2).}}
    \label{fig:vis_aug_image_1}
\vspace{-5pt}
\end{figure*}

\begin{figure*}[ht]
    \centering
    \includegraphics[width=1.0\textwidth]{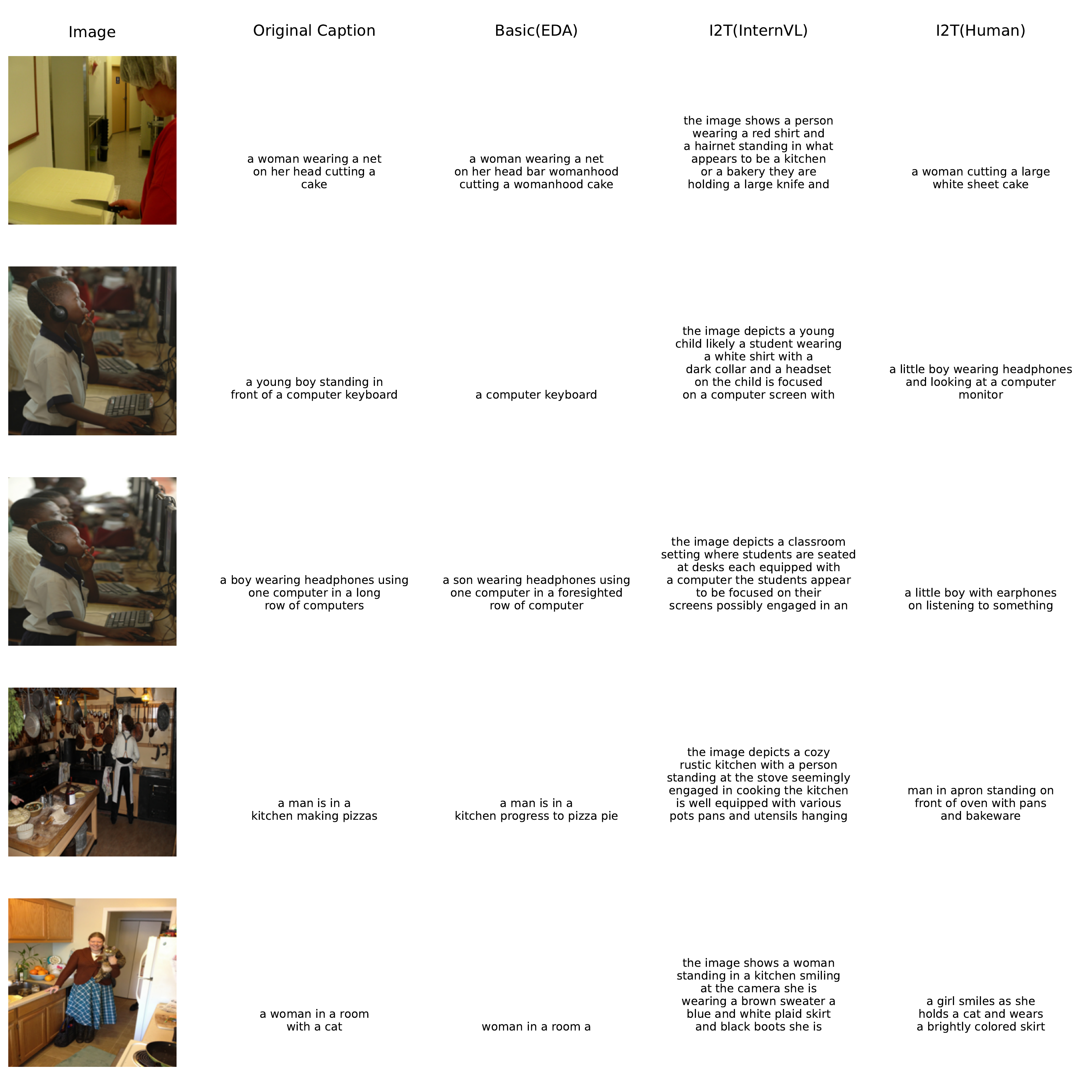}
    \vspace{-10pt}
    \caption{\textbf{Visualization of text augmentations (1).}}
    \label{fig:vis_aug_text_0}
\vspace{-5pt}
\end{figure*}

\begin{figure*}[ht]
    \centering
    \includegraphics[width=1.0\textwidth]{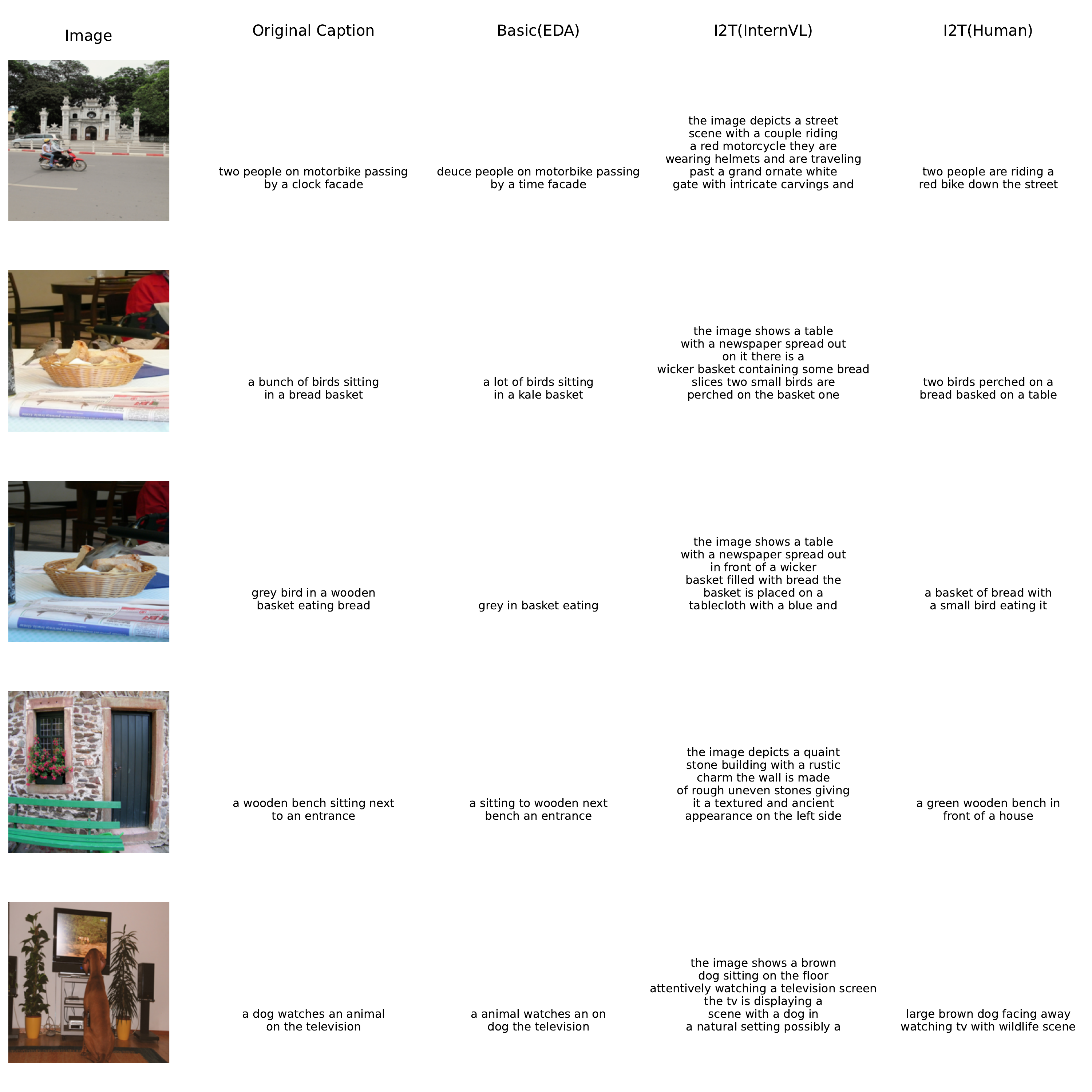}
    \vspace{-10pt}
    \caption{\textbf{Visualization of text augmentations (2).}}
    \label{fig:vis_aug_text_1}
\vspace{-5pt}
\end{figure*}

This section visualizes each augmentation technique.
Figure~\ref{fig:vis_aug_image_0} and Fig.~\ref{fig:vis_aug_image_1} show image augmentations, and Fig.~\ref{fig:vis_aug_text_0} and Fig.~\ref{fig:vis_aug_text_1} visualize text augmentations.
These visualizations help in understanding image-text alignment, augmentation diversity, and the distribution of augmentations.
For example, Fig.\ref{fig:vis_aug_image_0} and Fig.\ref{fig:vis_aug_image_1} show that while I2T(SD) generates diverse images, they differ significantly from the original images and often appear somewhat synthetic, potentially causing a distribution shift.
Additionally, Fig.~\ref{fig:vis_aug_text_0} and Fig.~\ref{fig:vis_aug_text_1} show that Basic(EDA), which is a basic word-level augmentation, can disrupt the image-text alignment.

\end{document}